\documentclass[letterpaper,journal]{IEEEtran}

\usepackage{amsmath,amsfonts}
\usepackage{array}
\usepackage{booktabs}
\usepackage{cite}
\usepackage{graphicx}
\usepackage{microtype}
\usepackage{multirow}
\usepackage{tabularx}
\usepackage{url}
\usepackage{float}
\usepackage{footnote}
\usepackage[hidelinks]{hyperref}

\newcolumntype{Y}{>{\raggedright\arraybackslash}X}
\newcommand{\ind}{\textsc{Independent}}
\newcommand{\ibsh}{\textsc{IBSH}}
\newcommand{\shared}{\textsc{Shared-Trunk}}
\begin{document}
\title{Ocular Recognition: Interoperability, Multi-Resolution, and Spectrum Challenges}



\author{Fabian~Stockhardt, Juan~E.~Tapia,~\IEEEmembership{Senior Member,~IEEE,}
and~Christoph~Busch,~\IEEEmembership{Fellow,~IEEE}%
\thanks{The authors are with the da/sec Biometrics and Internet Security Research Group, Hochschule Darmstadt, Germany (e-mail: \{juan.tapia-farias,christoph.busch\}@h-da.de).}%
\thanks{Draft manuscript, July 2026.}}

\maketitle

\begin{abstract}Ocular recognition has traditionally been developed under controlled acquisition conditions, which limits its applicability in semi-controlled and on-the-move biometric scenarios. A major obstacle is the lack of representative data covering the range of capture conditions encountered in practice. At a border-gate checkpoint, for example, ocular images acquired as a subject approaches the gate may vary substantially, from low-resolution visible-light periocular crops (RGB) to high-resolution near-infrared iris images (NIR). Effectively exploiting this heterogeneous evidence introduces three key design challenges: determining whether resolution-specific processing improves recognition across image resolutions; establishing a reliable way to compare templates produced by different resolution-specific processing routes; and allocating the available training data among specialists operating at different resolutions.

To investigate how resolution-specific processing and parameter sharing affect ocular recognition. We developed three AdaFace-based architecture families. The first comprises fixed-resolution specialists. The second uses the Independent Backbones with Shared Heads (\textit{IBSH}). The third introduces resolution-specific front-ends for $40\times40$, $80\times80$ and $160\times160$ inputs followed by a shared identity trunk and embedding projection. Each architecture is evaluated under route-specific resolutions, where each specialist receives its assigned training subset, and under full route, where every specialist receives the complete selected training pool.


The results identify a trade-off between same-resolution specialization, direct cross-resolution interoperability, and practical common-route re-extraction. Route-specific resolution produces pronounced route-dependent performance profiles, while full image pool route homogenizes the routes and primarily strengthens route $160\times160$. More training data per route therefore does not provide a uniform improvement.



\end{abstract}


\section{Introduction}
\label{sec:introduction}

\IEEEPARstart{F}{uture} biometric border-control systems may observe a traveler repeatedly while he approaches a gate for biometric on-the-move recognition. Currently, most border control systems are focused on cooperative faces and fingerprints because this information is included in the passport. However, fingerprint acquisition generally requires the traveler to present the hand in close proximity to a dedicated capture device. Other countries such as the UAE \footnote{\url{https://www.cl.cam.ac.uk/~jgd1000/UAEdeployment.pdf}} use iris recognition or a combination of at least two biometric modalities.

Recent border-management technology roadmaps, such as those from the European Border and Coast Guard Agency (Frontex), highlight "biometrics on the move" for future travel as an important operational topic \cite{frontex2023technology} using several complementary modalities captured at a distance, such as faces and periocular eye images.

\begin{figure}[H]
  \centering
  \includegraphics[scale=0.2]{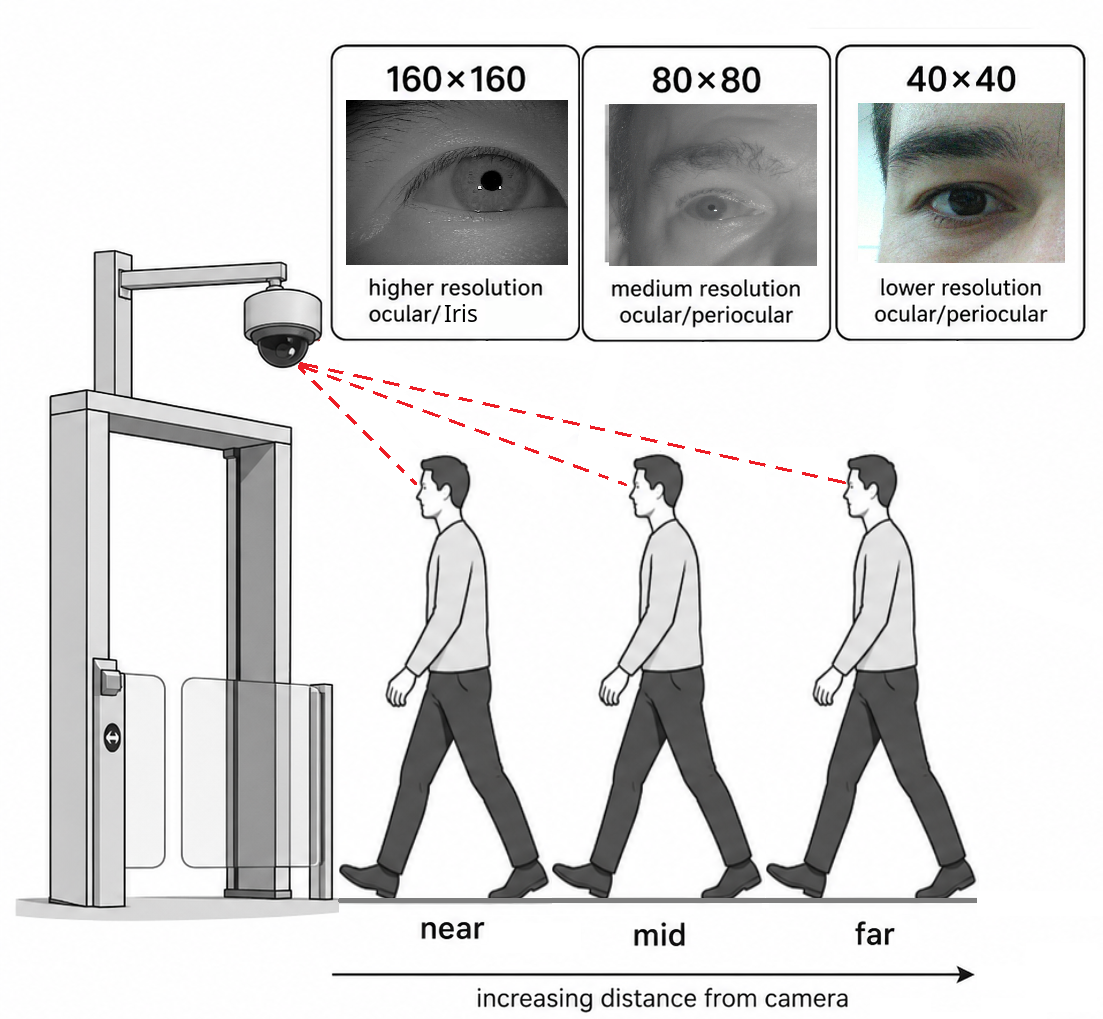}
  \caption{Conceptual iris-on-the-move scenario. Representative crops illustrate the evidence associated with the three evaluated network input sizes. They are not a temporal sequence of one individual. The labels $40/80/160$ denote resized tensors rather than native sensor resolutions (AI-generated image (ChatGPT Sol 5.6).}
  \label{fig:scenario}
\end{figure}

As distance decreases, the ocular region occupies more pixels. Motion, gaze, illumination, focus, sensor spectrum and challenging optical paths can still prevent controlled iris acquisition. A distant frame may contain only coarse periocular structure; a closer frame can reveal ocular context; only the most favorable observations may contain detailed iris texture. Rejecting every non-ideal frame discards potentially useful identity evidence. Unrelated recognizers also prevent the resulting templates from being compared directly. One conceptual example is illustrated in Figure \ref{fig:scenario}.

A severe data imbalance also compounds the challenge. Large face datasets supply millions of identity-labeled images, but their extracted eye regions are often small and low quality. Dedicated iris collections preserve considerably more local detail but contain far fewer identities and samples. Downsampling high-resolution images can discard fine identity-relevant detail, whereas upsampling low-resolution images produces redundant pixel values and can introduce interpolation artifacts. We investigate whether heterogeneous visible-spectrum (VIS) and near-infrared (NIR) periocular, ocular and iris-dominant observations can be combined within one recognition framework while retaining resolution-specific processing routes. This motivates comparing route-specific resolution, in which each specialist receives observations assigned to its resolution bin, with training on full image pools, in which every specialist receives the complete selected training pool.

The main research question is:

\begin{quote}
  \emph{How should resolution-specific processing, model coupling and training route be configured for ocular recognition across heterogeneous image resolutions?}
\end{quote}

We decompose it into three questions:




\textbf{RQ1}: What are the effects of resolution-specific modeling on same-resolution verification?

\textbf{RQ2}: How can verification be performed effectively across different route image resolutions?

\textbf{RQ3}: How does training-data routes influence the specialization and generalization of resolution-specific models?

Motivated by the previous questions, this work proposes the following main contributions:

\begin{itemize}
    \item It studies a coupling ladder comprising independent specialists, independent backbones with common class centers and a fixed-interface shared-trunk model, with common-route re-extraction as a practical baseline.

    \item It quantifies resolution-driven route specialization and the extent to which full route homogenizes performance across each routes.

    \item It evaluates same- and cross-resolution verification on five datasets reserved exclusively for testing.

\end{itemize}  

We study frame-level, single-eye templates and cosine distances for verification. Temporal aggregation, automatic routing, fusion with full face recognition and comparison with a controlled iris-code system remain outside the present experiments.

Section~\ref{sec:related-work} reviews related work. Section~\ref{sec:method} presents the method. Section~\ref{sec:metric} defines the evaluation metrics. Section~\ref{sec:protocol} describes the data and experimental protocol. Section~\ref{sec:results} reports the results. Section~\ref{sec:discussion} discusses their implications and limitations. Section~\ref{sec:conclusion} concludes the article.

\section{Related Work}
\label{sec:related-work}

Classical iris recognition combines highly discriminative texture with controlled image formation, segmentation, normalization and bit-wise comparison of binary iris codes~\cite{Daugman}. Periocular recognition can use the eyebrow, eyelids, lashes, sclera and surrounding skin area when the iris is insufficiently resolved, making it relevant to unconstrained acquisition~\cite{SharmaRoss2023PeriocularSurvey,AlonsoFernandez2024PeriocularUnconstrained}. Large-scale ocular cropping from face data has further shown that face collections can provide useful periocular identity supervision~\cite{AlonsoFernandez2025OcularCropping}. These observations motivate mixed training but do not determine how resolution-specific models should share parameters or training data. They also do not solve template compatibility between low-resolution VIS crops and detailed NIR or iris imagery.

Several applications from periocular eye images have been raised in recent years, focusing on periocular recognition in RGB \cite{Selfie-Tapia}, presentation attacks on the RGB \cite{Peri-PAD} or NIR spectrum \cite{tapia-pad-tifs, iris-livdet}, Digital attacks such as Iris Morph \cite{tapia-morph}, Explainability \cite{ExplainablePR} and many others.

Iris--periocular fusion addresses a related but distinct problem. Attention-based feature fusion~\cite{luo2021deepfusion} and quality-adaptive score fusion~\cite{yan2024boosting} combine simultaneously available iris and periocular evidence. Their component representations need not be directly comparable when only one region is available. Here, each sample follows one route and must independently produce a template that can be scored against a template from another route.

Cross-spectral ocular work has learned aligned latent representations or translated imagery into a common target spectrum~\cite{hernandezdiaz2020crossspectral}. Multi-resolution face recognition likewise uses resolution-specific branches and shared metric learning~\cite{zangeneh2017twobranch,wang2020resolution}. The Branch-to-Trunk Network (BTNet) provides an architectural reference for the shared-trunk family evaluated here. It attaches resolution-specific branches to levels of a pretrained face-recognition trunk and uses backward-compatible supervision and distillation~\cite{he2023btnet,shen2020backward}. 

AdaFace adapts its angular margin using the pre-normalization feature norm as a proxy for image recognizability~\cite{kim2022adaface}. This mechanism is relevant under heterogeneous resolution and unconstrained capture conditions. It does not by itself align independently parameterized embedding spaces or determine how heterogeneous training samples should be distributed among specialists.

Motivated by these gaps, we use adapted AdaFace as the common training objective. All the models are trained from scratch without a privileged teacher approach. Resolution also co-varies with spectrum, sensor, crop geometry and available anatomy.

\section{Method}
\label{sec:method}
This work evaluates two design axes within an imbalanced ocular training setting:  

1) The \textit{coupling axis} ranges from independent fixed-resolution models through parameter-disjoint specialists with shared class centers to resolution-specific front-ends followed by a shared high-level trunk. 

\begin{figure*}[t]
  \centering
  \includegraphics[width=0.99\textwidth]{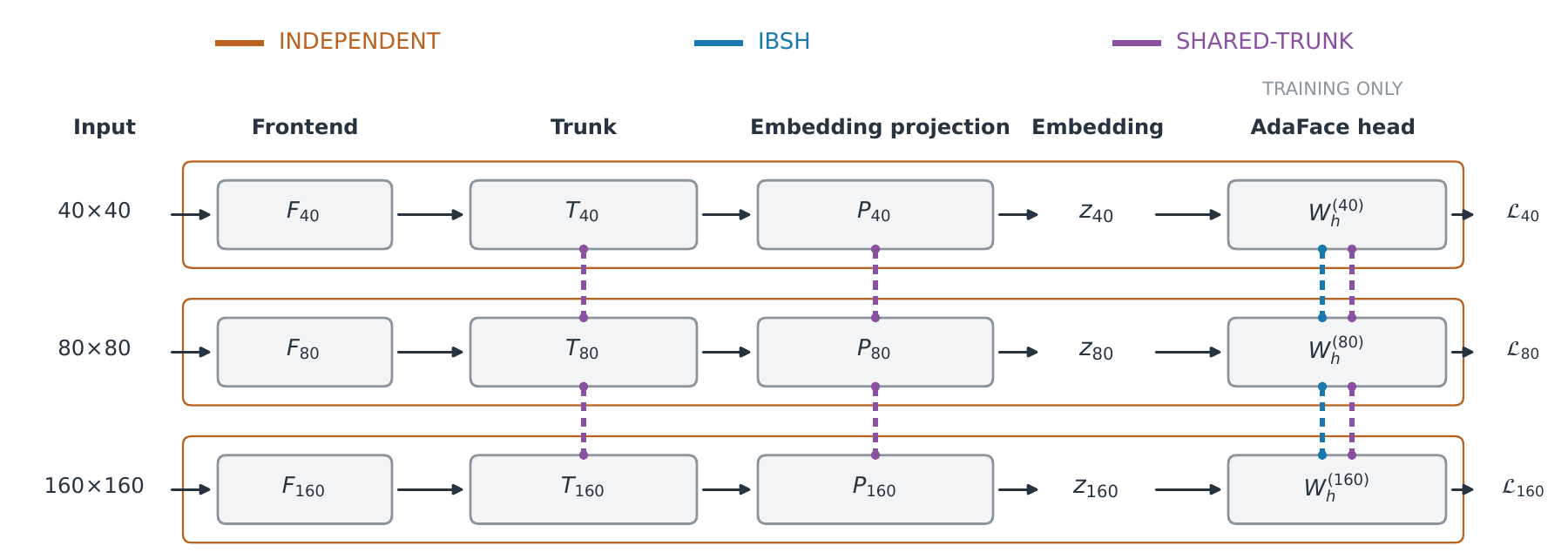}
    \caption{Architectural levels of resolution coupling. Each row represents a fixed input route. \textsc{Independent} uses separately optimized specialists with private frontends $F_r$, trunks $T_r$, embedding projections $P_r$, and dataset-specific AdaFace heads $W_h^{(r)}$. IBSH shares only the dataset-specific heads across routes, whereas \textsc{Shared-Trunk} additionally shares the trunk and embedding projection while retaining private frontends. Each set of vertically connected boxes represents route-specific applications of a single shared module; the colored links denote parameter sharing, not feature fusion. During training, the source dataset $h$ selects the active AdaFace head. During inference, the heads are removed, and verification uses cosine similarity between L2-normalized 512-D embeddings $z_r$.}
  \label{fig:architecture}
\end{figure*}

2) The \textit{resolution axis} compares multi-resolution training pools for every resolution route. Common-resolution re-extraction through an independent specialist provides a practical cross-resolution baseline. Figure \ref{fig:architecture} illustrates the entire process.


\subsection{Sample, branch and route definition}

Let $x$ be a single periocular eye crop \textit{\textbf{sample}} and let $y=(p,e)$ denote a subject $p$ and eye side $e$. Left and right eyes are different biometric classes. A \textit{\textbf{route}} ${r\in\ {40\times 40,80\times80, 160\times160}}$ produces a $(z)$ normalized 512-dimensional embedding:

\begin{equation}
  z=f(x,r)\in\mathbb{R}^{512}, \qquad \lVert z\rVert_2=1,
\end{equation}
and the biometric verification uses cosine similarity. Training data eye crops are assigned to resolution bin $(b)$ by their larger dimension according to each ($H\times W$):
\begin{equation}
 b(x)=\begin{cases}
 40\times40,&\max(H,W)<60,\\
 80\times80,&60\leq\max(H,W)<120,\\
 160\times160,&\max(H,W)\geq120.
 \end{cases}
\end{equation}
Square image samples are resized directly to the tensor size required by the target route. Rectangular samples are first padded to a square canvas and then resized. The \textit{\textbf{branch}} labels therefore denote fixed network input sizes rather than source image resolutions. 

\subsection{Coupling Axis}

An \textit{\ind{}} specialist is a complete fixed-resolution network with its own dataset-specific AdaFace heads. We optimize the three specialists separately ($40,80,160$), so no mechanism aligns their embedding spaces. To represent conventional fixed-input recognition, we use common-route re-extraction as a practical baseline. Both pre-resize source crops are rendered for either the lower route (lower-route rerouting) or the higher route (higher-route rerouting), then both embeddings are re-extracted with the corresponding independent specialist. 

These baselines use one embedding coordinate system and are not aligned for cross model comparison. The \textit{\ibsh{}} retains three complete parameter-disjoint networks but jointly trains them against shared dataset-specific class centers. The \textit{\shared{}} contains private resolution front-ends followed by one common residual trunk, projection and set of dataset heads. Table~\ref{tab:coupling} summarizes the three evaluated systems independently of their training pool of images.

\begin{table}[H]
  \centering
  \caption{Architectural levels of resolution coupling. Each system is evaluated under routed (1/3) and full (3/3) specialist resolutions.}
  \label{tab:coupling}
  \footnotesize
  \setlength{\tabcolsep}{4pt}
  \begin{tabular}{lccc}
    \toprule
    System & Joint training & Shared centers & Shared trunk\\
    \midrule
    \ind{} & no & no & no\\
    \ibsh{} & yes & yes & no\\
    \shared{} & yes & yes & yes\\
    \bottomrule
  \end{tabular}
\end{table}

\subsection{Resolution axis}
We evaluate each coupling level under two specialist-resolution routes. Under routed  (\textit{1/3}), source selection follows route-specific origin rules and samples are not systematically replicated across all routes. Higher-resolution observations may still be downscaled to fill a lower-resolution route according to the sampling protocol. Under full routes (\textit{3/3}), a common global source pool is selected, and every selected source is rendered for all three routes. The labels \textit{1/3} and \textit{3/3} describe the route rather than exact fractions of unique training images. The phase-specific sampling targets and realized resolution counts are reported in Section~\ref{sec:protocol}. Figure \ref{fig:architecture-detail} illustrates the different routes and resolutions.

\subsection{Independent backbones with shared heads (IBSH)}

Let $r=\{40,80,160\}$ denote the route set. For a source crop $x$, $x^{(r)}$ denotes the route-specific input tensor obtained by applying the padding and resizing procedure for route $r$. In \textit{\ibsh{}}, specialist $f_{\theta_r}$ produces a route-specific embedding vector $z_r$.
After L2 normalization, $z_r$ is the 512-D representation used for comparison and classification:

\begin{equation}
 z_r=f_{\theta_r}(x^{(r)})\in\mathbb{R}^{512},
 \qquad \lVert z_r\rVert_2=1.
\end{equation}
The trainable weights of the three specialists are disjoint: each route has its own convolutional backbone, batch-normalization affine parameters, PReLU parameters and final fully connected projection to the raw 512-D embedding before normalization.

We call one synchronized optimizer cycle a \emph{superstep}. Under route-specific image pools, a superstep packs at most one existing mini-batch per active route for the same dataset head. Samples need not correspond across routes. The active route set $\mathcal A_h\subseteq\mathcal R$ may contain one, two or three routes. Under training on the full image pool, one selected source batch is rendered at all three resolutions and the three aligned route batches form the superstep. 

The specialists evaluate their route-specific losses against the same dataset head $W_h$, then combine these losses before the synchronized parameter update. Let $\mathcal L_{h,r}$ denote the mean AdaFace classification loss over the $n_r$ retained samples processed by active route $r$. The sample-weighted superstep objective is:
\begin{equation}
 \mathcal{L}_{h}^{\mathrm{IBSH}}=
 \frac{\sum_{r\in\mathcal A_h}n_r\mathcal{L}_{h,r}}
 {\sum_{r\in\mathcal A_h}n_r}.
\end{equation}
This weighting gives every retained sample equal influence when route counts differ. To maintain one AdaFace quality calibration, the pre-normalization embedding norms from all active routes are detached and pooled so that the running quality statistics update once per superstep. With a per-route batch size of $512$, the shared head can observe up to $1,536$ route-specific views before an update. Each $\theta_r$ receives gradients only through its own active route, whereas $W_h$ receives contributions from every active route. These common class centers provide an alignment mechanism without mathematically guaranteeing that the independently parameterized embeddings share exact coordinates.

\subsection{Shared trunk}

The private frontends map their inputs to a common $20\times20\times128$ interface. Branch $40$ downsamples once, branch $80$ twice including its stem and branch $160$ three times. The shared residual suffix maps $20\times20\times128$ to $10\times10\times256$ and then $5\times5\times512$ ($E512$). Batch normalization, dropout ($p=0.4$), a linear $512-D$ projection, non-affine output normalization and $L_2$ normalization complete the embedding. All convolutional paths are initialized from scratch.

Unlike \textit{BTNet}, every route traverses exactly the same high-level mapping after the merge interface. The fixed merge makes shared processing explicit while allowing private frontends to preserve resolution-specific low-level evidence.

\begin{figure}[]
  \centering
  \includegraphics[scale=0.22]{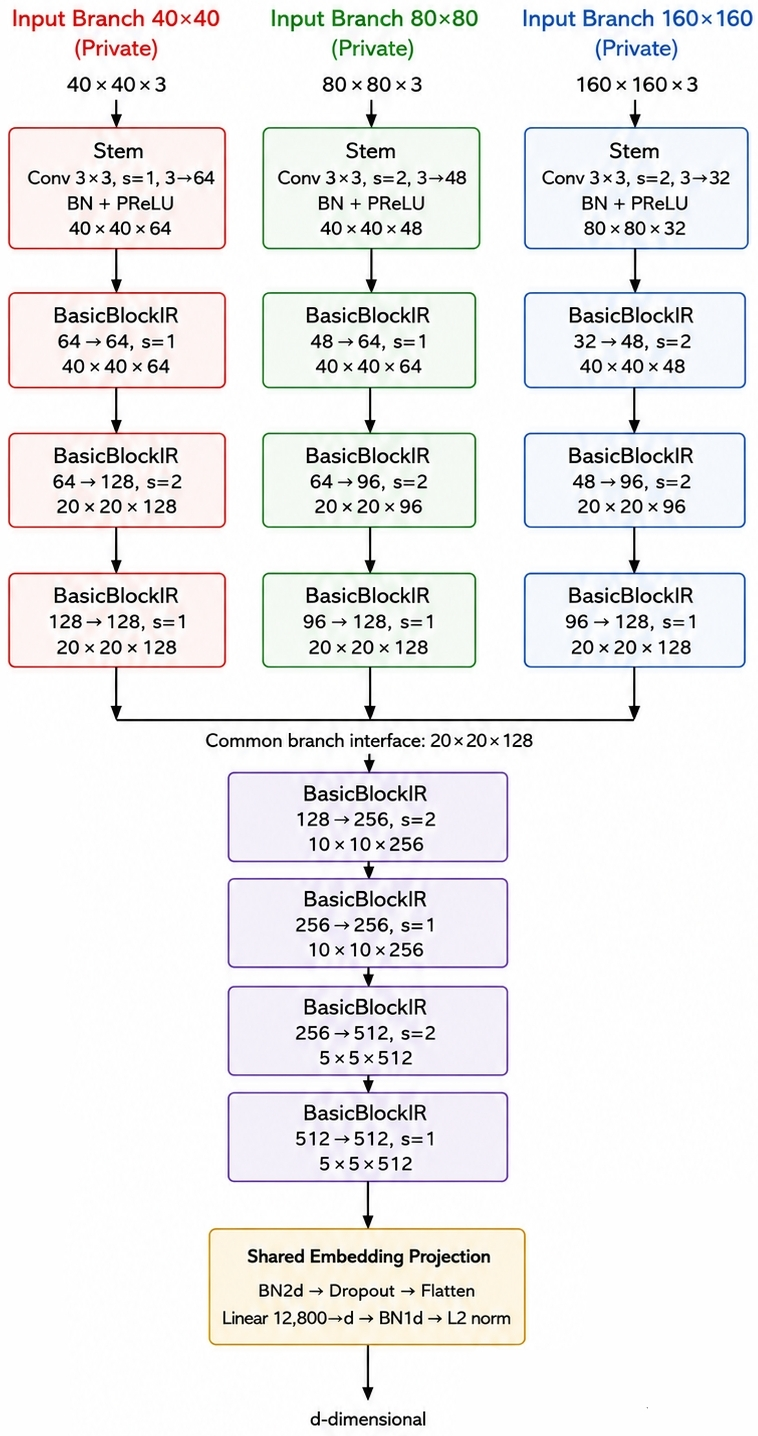}
  \caption{Shared-trunk model for the \textit{\textbf{inference}} process. The private frontends map their respective inputs to a common $20\times20\times128$ interface. The subsequent residual trunk and embedding projection are shared across routes. The symbol $d$ denotes the configured embedding dimension. Dataset-specific AdaFace heads are omitted from end of the figure.}
  \label{fig:architecture-detail}
\end{figure}

Figure~\ref{fig:architecture} illustrates the end-to-end system context, while Figure~\ref{fig:architecture-detail} specifies the layer-level branch and trunk configuration.

\subsection{Dataset-specific AdaFace heads}

Identity annotations are dataset-local and are not reconciled across independently collected datasets. Internet-scraped data and different~\textit{CASIA} dataset collections may contain the same real-world subject under unrelated identifiers. Following the dataset-aware classification principle of~DAIL~\cite{wang2021dail}, each evaluated architecture uses dataset-specific \textit{AdaFace} classifiers rather than a single, global classification head spanning all dataset-specific identity classes~\cite{kim2022adaface}. Their sharing across resolution routes depends on the coupling level: every \ind{} specialist owns a separate classifier set, whereas \ibsh{} and \shared{} use one classifier per dataset across routes. For a sample from dataset $h$, only the active classifier's centers $W_h$ participate in the cross-entropy loss (CE); in the independent baseline, $W_h$ denotes the specialist-local copy:

\begin{equation}
 \mathcal{L}_h=\operatorname{CE}\!\left(
 \operatorname{AdaFace}_h(z,\lVert \tilde z\rVert_2),y_h\right).
\end{equation}

VIS and NIR images are expected to have consistent identities within one dataset and therefore remain in the same head. Separate heads isolate dataset-specific label spaces. 

Such isolation is particularly important for cross-trait training. The same real-world subject may, for example, participate in both face and iris acquisition sessions, including sessions distributed across different \textit{CASIA} collections, without receiving a persistent cross-dataset or cross-trait identifier. Combining dataset-local identity labels from all sources into a single classification head could assign two records of that subject to distinct, competing classes. Dataset-specific heads avoid directly enforcing separation between such cross-dataset records because classes from different datasets never compete within the same softmax.

Dataset-specific heads also provide a modular path for expanding the training registry. A future \textit{VIS--NIR}, periocular, or iris dataset can receive a new head and contribute classification gradients to the participating feature network or networks without first resolving all of its identities against every existing source. The design reduces the cross-dataset identity-curation burden. It does not eliminate label errors within a dataset, align unknown duplicate identities or provide continual learning without joint retraining and appropriate source balancing.

\subsection{Parameter reuse and active-path computation}

The three independent specialists and \textit{IBSH} contain the same three complete route networks and require $52.68M$ parameters as a library. Reusing the trunk and projection reduces the \shared{} library to $18.58M$ parameters, which represent a $64.73$\% reduction. The active path itself is structurally unchanged. The benefit concerns model storage rather than inference speed. Table~\ref{tab:compute} shows inference parameters and convolution and multiply--accumulate operations (MACs). 

MACs omit normalization, activation, pooling, routing, cosine scoring and the detection/alignment pipeline. Learned normalization parameters are counted, whereas running-statistic buffers are not.
The six dataset-specific AdaFace classification heads add $8.877M$ trainable parameters during training. 

\begin{table}[H]
  \centering
  \caption{\textit{E512} inference-model. Active parameters and \textit{Conv+Linear} GMACs are identical for \ind{}, \ibsh{} and \shared{}.}
  \label{tab:compute}
  \scriptsize
  \setlength{\tabcolsep}{3.4pt}
  \begin{tabular}{lrrr}
    \toprule
    Quantity & Route 40 & Route 80 & Route 160\\
    \midrule
    Active parameters [M] & 17.655 & 17.539 & 17.491\\
    Conv+Linear GMAC/image & 1.022 & 0.945 & 0.942\\
    \midrule
    \multicolumn{2}{l}{Stored three-route library} & \multicolumn{2}{r}{\ind{}/\ibsh{}: 52.684M}\\
    \multicolumn{2}{l}{Stored shared library} & \multicolumn{2}{r}{\shared{}: 18.580M}\\
    \bottomrule
  \end{tabular}
\end{table}

\begin{figure*}[t]
  \centering
  \setlength{\tabcolsep}{3pt}
  \renewcommand{\arraystretch}{1.05}
  \newcommand{\datasetexample}[2]{%
    \begin{tabular}{@{}c@{}}
      \includegraphics[height=0.80in,keepaspectratio]{#2}\\[-2pt]
      \scriptsize #1
    \end{tabular}}
  \footnotesize Training datasets\\[2pt]
  \begin{tabular}{@{}cccccc@{}}
    \datasetexample{VGGFace2}{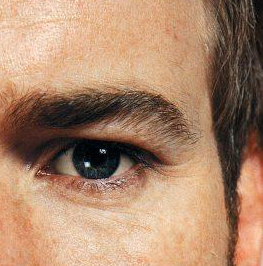} &
    \datasetexample{CASIA NIR--VIS}{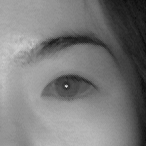} &
    \datasetexample{ND-NIVL}{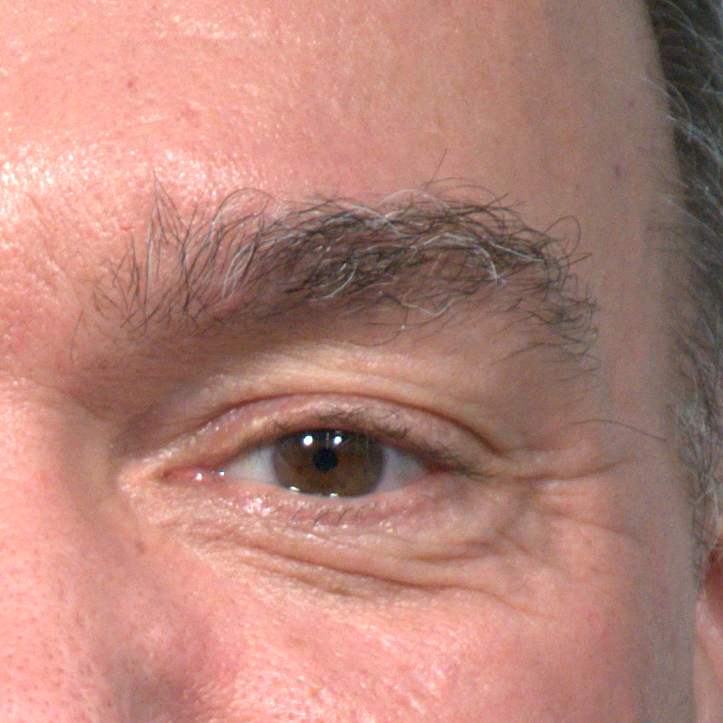} &
    \datasetexample{PolyU}{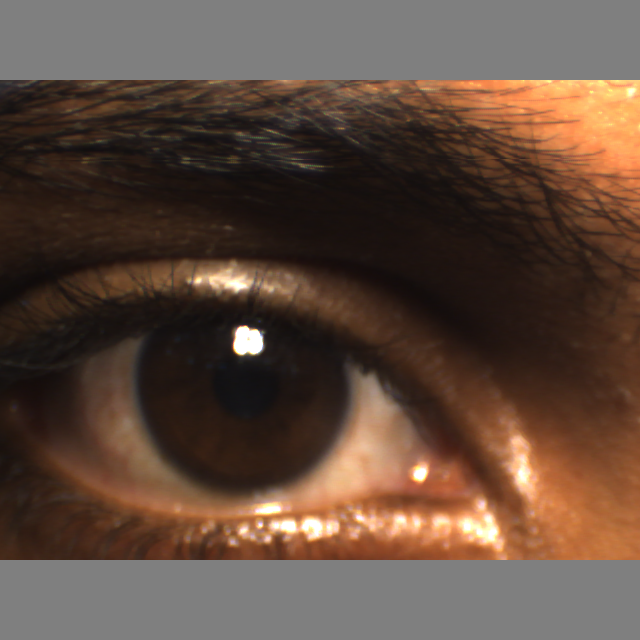} &
    \datasetexample{CASIA-Lamp}{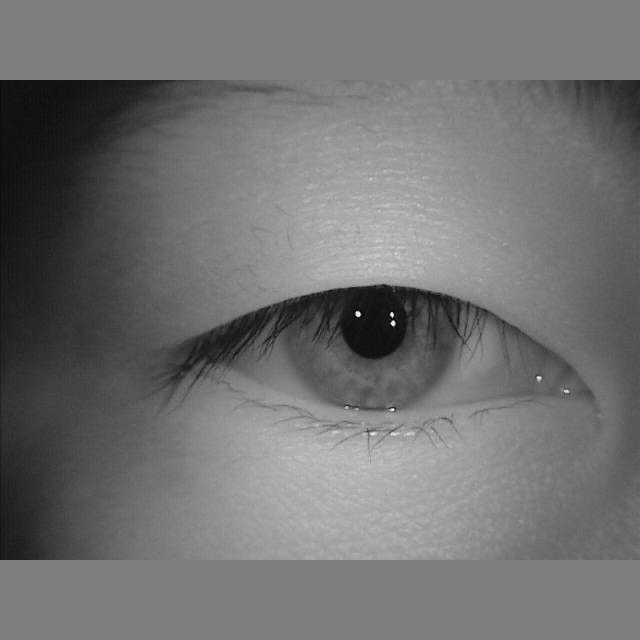} &
    \datasetexample{CASIA-Thousand}{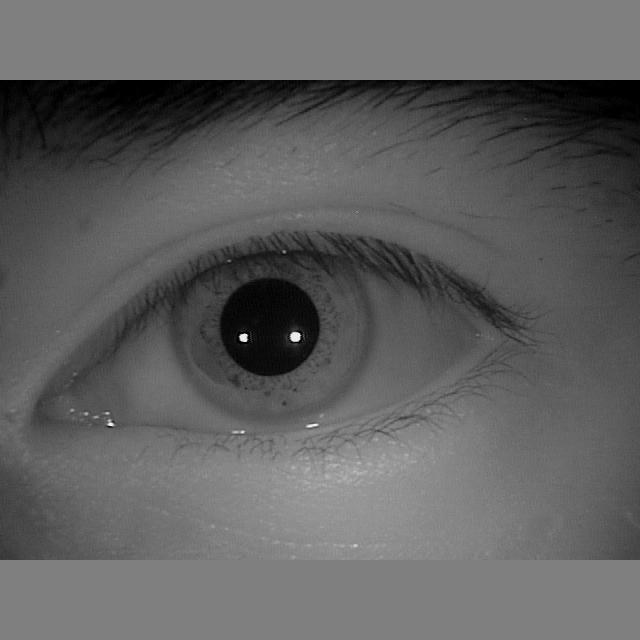}
  \end{tabular}\\[4pt]
  \footnotesize Test datasets\\[2pt]
  \begin{tabular}{@{}ccccc@{}}
    \datasetexample{FERET}{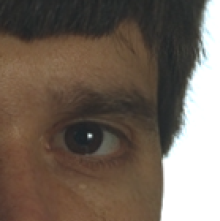} &
    \datasetexample{UFPR}{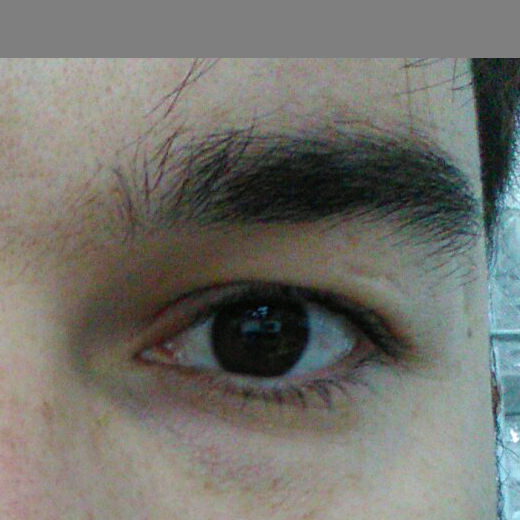} &
    \datasetexample{MCXFace}{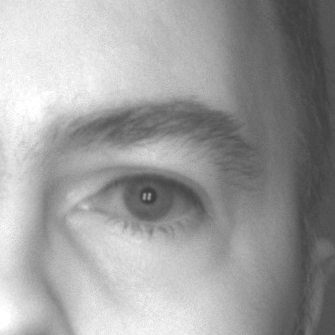} &
    \datasetexample{IIT Delhi}{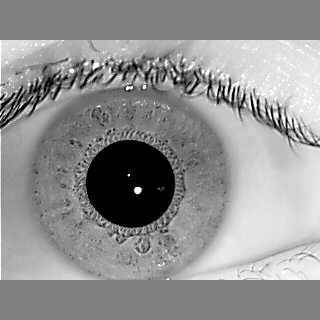} &
    \datasetexample{MMU1}{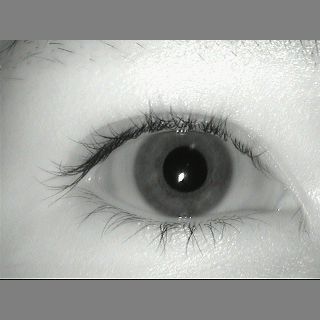}
  \end{tabular}
  \caption{Representative pre-resize network crops from all training and source-disjoint test datasets, shown before branch-specific resizing.}
  \label{fig:training-examples}
\end{figure*}

\section{Metrics}
\label{sec:metric}

In order to compare the biometric performance of all models, we use standard verification metrics based on mated and non-mated comparisons. Each subject-eye combination is treated as a distinct biometric identity. Mated comparisons therefore involve the same subject and eye side, whereas non-mated comparisons involve different subjects. 

Verification is based on the cosine similarity $(s)$ between two L2-normalized embeddings $\mathbf{z}_i$ and $\mathbf{z}_j$,
\begin{equation}
\label{eq:cosine_similarity}
    s(\mathbf{z}_i,\mathbf{z}_j)
    = \mathbf{z}_i^{\top}\mathbf{z}_j ,
\end{equation}
where larger scores indicate greater similarity. Given a decision threshold $\tau$, a verification attempt is accepted if $s \geq \tau$ and rejected otherwise.

The \emph{false non-match rate} (FNMR) at threshold $\tau$ is defined:
\begin{equation}
\label{eq:FNMR}
    \mathrm{FNMR}(\tau)
    =
    \frac{
        \left|\{\text{mated comparisons with } s < \tau\}\right|
    }{
        \left|\{\text{all mated comparisons}\}\right|
    },
\end{equation}
that is, the proportion of mated comparisons that are incorrectly rejected. Conversely, the \emph{false match rate} (FMR) at threshold $\tau$ is defined as:
\begin{equation}
\label{eq:FMR}
    \mathrm{FMR}(\tau)
    =
    \frac{
        \left|\{\text{non-mated comparisons with } s \geq \tau\}\right|
    }{
        \left|\{\text{all non-mated comparisons}\}\right|
    },
\end{equation}
that is, the proportion of non-mated comparisons that are incorrectly accepted.

Biometric performance is reported using the \emph{Equal Error Rate} (EER) and the \emph{False Non-match Rate at a fixed False Match Rate} (FNMR@FMR). The EER is the error rate at the operating point where FMR and FNMR are equal. Lower EER indicates better verification performance.

The fixed operating point used throughout this study is FNMR at $\mathrm{FMR}=1\times10^{-3}$. It quantifies the proportion of mated comparisons rejected at the decision threshold. Lower FNMR@FMR indicates better verification performance.

\section{Data and Experimental Protocol}
\label{sec:protocol}

\subsection{Training datasets}
The training data comprise six datasets grouped into three broad categories: face-derived VIS, face-derived cross-spectral VIS/NIR and NIR iris-native samples.

The \textit{CASIA NIR-VIS 2.0}~\cite{li2013casia} database offers pairs of mugshot images and their corresponding Near InfraRed (NIR) and Visible-Spectrum~(VIS) images. The images in this database were collected in four recording sessions, with the first session identical to the~\textit{CASIA-HFB} database \cite{HFB}. It consists of 725 subjects in total. There are [1-22] VIS and [5-50] NIR face images per subject. The eye positions are also distributed with the images.

The \textit{ND-NIVL}~\cite{ND-NIVL} dataset was collected by the University of Notre Dame, USA. The ND-NIVL contains VIS and NIR face images from the same subjects. The capture process was carried out over two semesters (fall 2011 and spring 2012). The VIS images were collected using a \textit{Nikon-D90} camera. The \textit{Nikon-D90} uses a $23.6\times15.8$ mm CMOS sensor; the resulting images have a resolution of $4288\times2848$ pixels. The images were acquired using automatic focus settings. All images were acquired under normal indoor lighting at a 5-foot stand-off with a frontal pose and neutral expression.

The \textit{PolyU Cross-Spectral Iris Images Database}~\cite{polyu_cross_iris} contains $12,540$ iris images from $209$ subjects. Both eyes were captured simultaneously in VIS and NIR, with 15 images per spectrum and eye at $640\times480$ pixels.

\textit{CASIA-Iris-Lamp} and \textit{CASIA-Iris-Thousand}~\cite{casia_irisv4_website} are NIR iris subsets of CASIA-IrisV4. The \textit{CASIA-Iris-Thousand} contains $20,000$ images from $1,000$ subjects acquired with the dual-eye \textit{IKEMB-100} iris camera. Its principal within-class variations include eyeglasses and specular reflections.

\textit{VGGFace2}~\cite{VGGFACE} is a large-scale VIS face dataset collected from web imagery and designed to cover substantial variation in pose, age, illumination, ethnicity as well as profession. The original dataset contains 3.31 million images of 9,131 identities, with an official training split of 8,631 identities and a test split of 500 identities.

In this work, VGGFace2 is not used as a face-recognition benchmark. Instead, it provides the large-scale VIS source from which face-derived eye crops are extracted. Because non-$112{\times}112$-aligned source images and crop metadata were available to us, \textit{VGGFace2} can populate all three resolution routes. The actual training manifest used by the sampler contains 3,579,290 \textit{VGGFace2} eye-crop images. The remaining training sources contribute 29,060 \textit{CASIA NIR--VIS}, 39,410 \textit{ND-NIVL}, 9,480 PolyU, 13,330 \textit{CASIA-Lamp} and 15,860 \textit{CASIA-Thousand} images. The Figure~\ref{fig:training-examples} illustrates representative observations from the heterogeneous training sources.

\subsection{Experimental Protocol}

Every dataset used for model development has a fixed 80/10/10 person-disjoint train/validation/test split constructed according to the provided identity annotations. Results on the internal 10\% test partitions are omitted in favor of the source-dataset-disjoint evaluation described in the next subsection.

The training sources differ substantially in size, spectral modality and native crop resolution. A naive use of the available pools would produce an imbalanced multi-route training signal dominated by \textit{VGGFace2}. We therefore define two related exposure protocols over the fixed 50-epoch schedule. The \emph{routed} (1/3) protocol assigns each selected sample to one route according to its resolution bin and the configured route fractions. The \emph{full} (3/3) protocol selects a common source image pool and presents every selected sample to all three routes after route-specific resizing.

Neither protocol makes all datasets equally large. Non-\textit{VGGFace2} sources are sampled completely from their available training pools. \textit{VGGFace2} is used as the large VIS fill source while its contribution is capped to prevent it from overwhelming the dataset-head mixture. Sampling is repeated every epoch, so Tables~\ref{tab:primary-routing} and~\ref{tab:exposure-regimes} report accumulated frequency of training on the full image pool rather than unique images.

\begin{table}[H]
  \centering
  \caption{Mean realized image routed (1/3) over the complete 50-epoch schedule and three seeds. Counts accumulate repeated sampling across epochs and do not denote unique images. Route percentages give the within-dataset distribution of image pools. Class counts use subject-eye labels.}
  \label{tab:primary-routing}
  \scriptsize
  \begin{tabular}{@{}lrrrrr@{}}
    \toprule
    Dataset & Classes & Image pools & 40 [\%] & 80 [\%] & 160 [\%] \\
    \midrule
    VGGFace2        & 12,502 & 21,835,710 & 41.38 & 38.91 & 19.71 \\
    CASIA NIR--VIS  & 1,198  & 1,792,200  & 20.58 & 45.13 & 34.28 \\
    ND-NIVL         & 1,062  & 2,031,718  & 0.00  & 3.01  & 96.99 \\
    PolyU           & 316    & 488,803    & 0.00  & 3.03  & 96.97 \\
    CASIA-Lamp      & 673    & 680,960    & 0.00  & 2.26  & 97.74 \\
    CASIA-Thousand  & 1,586  & 818,103    & 0.00  & 3.07  & 96.93 \\
    \bottomrule
  \end{tabular}
\end{table}

Routed image pools follows three phases: a 5-epoch warm-up, a 30-epoch mixed phase and a 15-epoch fine-tuning phase. The target route fractions $(40,80,160)$ are $(0.40,0.35,0.25)$, $(0.35,0.35,0.30)$ and $(0.25,0.30,0.45)$ for the three phases, respectively. The corresponding \textit{VGGFace2} target fractions per route are $(0.97,0.95,0.60)$, $(0.96,0.90,0.50)$ and $(0.95,0.85,0.40)$. Later phases therefore increase the relative contribution of the route-160 while reducing the \textit{VGGFace2} contribution, particularly for the 80 and 160 routes. 

Under training on the full image pool, a global source pool is formed with phase-specific \textit{VGGFace2} target fractions of $87.05\%$, $80.10\%$ and $67.25\%$. These fractions are derived from the route-weighted targets of the routed schedule. Every selected source, including \textit{VGGFace2} samples from all native-resolution bin, is subsequently processed into one input tensor for each route using crop, padding and route-specific resizing.


Table~\ref{tab:exposure-regimes} reports the number of samples received by each route at each resolution for each route-specific network.
For \shared{}, the combined value is the number of route-specific feature views processed by the common trunk and shared classification heads. For \ibsh{}, it is the number of sample contributions received by the shared classification heads across all route backbones. For \ind{}, the combined value is only the sum over three separately trained models; no component jointly observes these image pools. The counts characterize the complete training schedule rather than cumulative iamge pools at the validation-selected checkpoint.

\begin{table}[H]
  \centering
  \caption{Mean training on route-specific image pools over the complete 50-epoch schedules and three seeds. Values are millions of image views.}
  \label{tab:exposure-regimes}
  \scriptsize
  \setlength{\tabcolsep}{4.5pt}
  \begin{tabular}{lrrrr}
    \toprule
    Image pool regime & Route 40 & Route 80 & Route 160 & Combined \\
    \midrule
    Routed (1/3) & 9.405 & 9.422 & 8.821 & 27.647 \\
    Full (3/3) & 25.190 & 25.190 & 25.190 & 75.571 \\
    \bottomrule
  \end{tabular}
\end{table}

Under training on route-specific image pools, \textit{VGGFace2} and \textit{CASIA-NIR--VIS} contribute substantially to all three routes, whereas ND-NIVL and the iris-dominant datasets are routed predominantly through the 160 branch. The small route-80 contributions of these sources result from the sampling and fill procedure and do not indicate a corresponding route-80 subset based on their source image resolutions.

After route selection, square images are resized directly to the input size required by the selected branch, whereas rectangular images are first padded to a square canvas and then resized. Validation and testing use no stochastic crop changes.

\subsection{Dataset-disjoint test protocols}

FERET~\cite{phillips1998feret} and UFPR-Periocular~\cite{zanlorensi2022new} test VIS face-derived and periocular observations; MCXFace~\cite{george2022prepended} is evaluated with separate VIS--VIS, VIS--NIR and NIR--NIR test protocols; IIT Delhi~\cite{iitd_iris} and MMU1~\cite{mmu_iris} test iris-oriented transfer. The Figure~\ref{fig:training-examples} illustrates representative observations from the test sources.

Consistent with the metric definitions, all test protocols are evaluated side-specifically: left-eye samples are compared only with left-eye samples and right-eye samples only with right-eye samples. The main evaluation uses only source datasets excluded entirely from training and validation (Table~\ref{tab:test-data}). 

\begin{table}[H]
  \centering
  \caption{Dataset-disjoint test protocols used in this paper. MCXFace counts sum its VIS--VIS, VIS--NIR and NIR--NIR test-protocol variants.}
  \label{tab:test-data}
  \footnotesize
  \setlength{\tabcolsep}{8pt}
  \begin{tabular}{@{}llrr@{}}
    \toprule
    Dataset & Primary domain & Mated & Non-mated\\
    \midrule
    FERET & VIS face-derived ocular & 7,284 & 5,902,017\\
    UFPR-Periocular & VIS periocular & 235,222 & 50,000,000\\
    MCXFace & VIS/NIR ocular & 51,768 & 1,749,568\\
    IIT Delhi Iris & NIR iris-dominant & 4,805 & 999,040\\
    MMU1 Iris & iris-dominant & 900 & 39,600\\
    \bottomrule
  \end{tabular}
\end{table}

\subsection{Training and selection configuration}

All the training models use 512-D embeddings and a per-route batch size of $512$. AdaFace model uses the following parameters: $m=0.4$, $h=0.333$, $s=64$ and $t_\alpha=0.01$. Stochastic gradient descent uses an initial learning rate of $0.1$, momentum $0.9$ and weight decay $10^{-4}$; the learning rate drops at epochs $20$, $34$ and $44$. Each principal experiment is repeated with three random seeds, $42$, $44$ and $46$. Each run contains $50$ epochs. Unless stated otherwise, reported test metrics are the mean and standard deviation across these three runs. Model selection is performed independently for each seed using the fixed validation protocol.

Checkpoint selection uses only the validation partitions of the 80/10/10 split. For each system, we select the checkpoint with the best unweighted macro score over within-branch, cross-branch, \textit{VIS--NIR} and iris validation protocols. Test-only datasets are never used for model selection.

Aggregation is performed hierarchically without weighting by pair count. For each seed and route, the three \textit{MCXFace} spectral test-protocol variants are first averaged to one \textit{MCXFace} value. Route-level summaries then use an unweighted arithmetic mean over the five test-only datasets. 

Same-resolution and cross-resolution summaries use unweighted means over their three respective routes. The reported uncertainty is the sample standard deviation across seeds. Scores are not pooled between datasets. Same-resolution denotes the three diagonal routes $40\rightarrow40$, $80\rightarrow80$ and $160\rightarrow160$. 

Cross-resolution denotes the three evaluated upper-triangular route pairs \(40\rightarrow80\), \(40\rightarrow160\) and \(80\rightarrow160\), where the first route processes the first image of each fixed comparison pair and the second route processes the second image.

Every system uses the same fixed pair definitions and pair counts. Equal weighting prevents the very large \textit{UFPR} test protocol from dominating the summary.

Within each exposure regime, system comparisons use the same source-selection schedule and route targets. They are not compared by optimizer-step count or wall-clock compute. The \ind{} specialists are optimized separately, whereas \ibsh{} can execute up to three full backbones within one superstep. The \ibsh{}-to-\shared{} comparison therefore remains a system comparison rather than an isolated trunk ablation.

\section{Results}
\label{sec:results}
This section summarizes the findings obtained from the proposed approaches and discusses them in relation to the research questions. Table~\ref{tab:same-eer} presents the performance comparison among the different routes, while Table~\ref{tab:cross-eer} reports the results of the corresponding cross-resolution strategies. Together, these results provide the basis for evaluating the effectiveness of the proposed routing schemes and drawing the main conclusions of this study.

\subsection{RQ1:What are the effects of resolution-specific modeling on same-resolution verification?}

Table~\ref{tab:same-eer} reports every same-resolution dataset route condition. The all-condition macro provides an average, but the route-level results show that no configuration is uniformly strongest. 

Training on full image pools \shared{} obtains the lowest all-condition \textit{EER} at $6.202{\pm}0.870$\%, whereas routed \shared{} obtains the lowest \textit{FNMR@FMR} at $26.333{\pm}2.842$\%.

Routed \shared{} is strongest on the route-40 macro for both metrics. At route-80, it has the lowest \textit{EER}, while routed \ind{} has the slightly lower \textit{FNMR@FMR}. Full image pools route \shared{} is strongest at route-160 for both metrics. The dataset route cells further quantify these averages. Routed \ind{} is particularly strong for \textit{FERET}, \textit{UFPR} and \textit{MCXFace} at routes-40 or routes-80. 

The training on full image pools \shared{} is strongest across all three \textit{IIT Delhi} routes, whereas \textit{MMU1} frequently favors the full-route independent specialists.

These results demonstrate route-dependent specialization under the routed image pool but do not identify a universally optimal coupling strategy. The direct effect of routed relative to train on the full image pool is examined separately in RQ3.

\begin{table*}[t]
  \centering
  \caption{Same-resolution performance (\%) for all dataset route conditions. Each cell reports \textit{EER} (top) and \textit{FNMR@FMR} at $1\times10^{-3}$ (bottom) as mean $\pm$ sample standard deviation over seeds $42$, $44$ and $46$. \textit{MCXFace} spectral test protocols are averaged within each route and seed. Bold values are row-wise best for each metric.}
  \label{tab:same-eer}
  \scriptsize
  \setlength{\tabcolsep}{1.5pt}
  \renewcommand{\arraystretch}{1.0}
  \begin{tabular}{@{}ll@{\hspace{2.5pt}\vrule width 0.2pt\hspace{2.5pt}}cc@{\hspace{2.5pt}\vrule width 0.2pt\hspace{2.5pt}}cc@{\hspace{2.5pt}\vrule width 0.2pt\hspace{2.5pt}}cc@{}}
    \toprule
    & & \multicolumn{2}{c}{\ind{}} & \multicolumn{2}{c}{\ibsh{}} & \multicolumn{2}{c}{\shared{}}\\
    \cmidrule(lr){3-4}\cmidrule(lr){5-6}\cmidrule(lr){7-8}
    Dataset & Route & 1/3 & 3/3 & 1/3 & 3/3 & 1/3 & 3/3\\
    \midrule
    \raisebox{3.6pt}{FERET} & \raisebox{3.6pt}{$40\!\rightarrow\!40$} & \shortstack{$\mathbf{1.30{\pm}0.06}$\\[-1.4pt]$6.81{\pm}0.65$} & \shortstack{$5.50{\pm}3.57$\\[-1.4pt]$30.29{\pm}16.43$} & \shortstack{$2.05{\pm}0.62$\\[-1.4pt]$20.84{\pm}18.01$} & \shortstack{$4.61{\pm}3.91$\\[-1.4pt]$36.76{\pm}34.00$} & \shortstack{$1.42{\pm}0.08$\\[-1.4pt]$\mathbf{6.34{\pm}1.11}$} & \shortstack{$4.32{\pm}2.52$\\[-1.5pt]$22.37{\pm}12.38$}\\[1.4pt]
    \raisebox{3.6pt}{FERET} & \raisebox{3.6pt}{$80\!\rightarrow\!80$} & \shortstack{$\mathbf{1.18{\pm}0.19}$\\[-1.4pt]$\mathbf{5.30{\pm}0.87}$} & \shortstack{$7.11{\pm}2.35$\\[-1.5pt]$53.46{\pm}12.12$} & \shortstack{$1.56{\pm}0.20$\\[-1.5pt]$8.44{\pm}1.74$} & \shortstack{$5.68{\pm}6.17$\\[-1.5pt]$29.80{\pm}31.27$} & \shortstack{$2.04{\pm}0.34$\\[-1.5pt]$10.83{\pm}2.15$} & \shortstack{$5.09{\pm}2.41$\\[-1.5pt]$28.47{\pm}13.61$}\\[1.4pt]
    \raisebox{3.6pt}{FERET} & \raisebox{3.6pt}{$160\!\rightarrow\!160$} & \shortstack{$19.72{\pm}3.73$\\[-1.5pt]$70.52{\pm}6.01$} & \shortstack{$8.47{\pm}1.24$\\[-1.5pt]$58.59{\pm}8.03$} & \shortstack{$23.39{\pm}3.89$\\[-1.5pt]$72.76{\pm}4.42$} & \shortstack{$\mathbf{3.79{\pm}3.04}$\\[-1.5pt]$\mathbf{23.93{\pm}18.41}$} & \shortstack{$11.54{\pm}5.17$\\[-1.5pt]$54.12{\pm}17.96$} & \shortstack{$5.62{\pm}2.31$\\[-1.5pt]$28.91{\pm}8.04$}\\[1.5pt]
    \cmidrule(lr){1-8}
    \raisebox{3.6pt}{UFPR} & \raisebox{3.6pt}{$40\!\rightarrow\!40$} & \shortstack{$6.69{\pm}0.01$\\[-1.5pt]$32.32{\pm}0.51$} & \shortstack{$7.70{\pm}1.29$\\[-1.5pt]$37.59{\pm}5.72$} & \shortstack{$7.55{\pm}0.62$\\[-1.5pt]$38.67{\pm}2.13$} & \shortstack{$8.02{\pm}0.84$\\[-1.5pt]$50.30{\pm}14.70$} & \shortstack{$\mathbf{6.01{\pm}0.31}$\\[-1.5pt]$\mathbf{28.53{\pm}1.53}$} & \shortstack{$7.90{\pm}1.14$\\[-1.5pt]$37.09{\pm}5.19$}\\[1.5pt]
    \raisebox{3.6pt}{UFPR} & \raisebox{3.6pt}{$80\!\rightarrow\!80$} & \shortstack{$\mathbf{5.75{\pm}0.23}$\\[-1.5pt]$\mathbf{27.77{\pm}1.43}$} & \shortstack{$8.19{\pm}0.77$\\[-1.5pt]$38.41{\pm}3.05$} & \shortstack{$6.41{\pm}0.15$\\[-1.5pt]$34.16{\pm}2.48$} & \shortstack{$9.07{\pm}0.78$\\[-1.5pt]$57.20{\pm}14.27$} & \shortstack{$5.93{\pm}0.30$\\[-1.5pt]$28.88{\pm}0.92$} & \shortstack{$7.85{\pm}0.69$\\[-1.5pt]$38.73{\pm}3.56$}\\[1.5pt]
    \raisebox{3.6pt}{UFPR} & \raisebox{3.6pt}{$160\!\rightarrow\!160$} & \shortstack{$17.42{\pm}3.06$\\[-1.5pt]$72.58{\pm}11.64$} & \shortstack{$9.26{\pm}0.12$\\[-1.5pt]$60.40{\pm}9.20$} & \shortstack{$20.81{\pm}6.95$\\[-1.5pt]$81.07{\pm}13.74$} & \shortstack{$\mathbf{7.87{\pm}0.59}$\\[-1.5pt]$43.35{\pm}5.15$} & \shortstack{$10.01{\pm}1.33$\\[-1.5pt]$50.74{\pm}8.89$} & \shortstack{$8.39{\pm}0.89$\\[-1.5pt]$\mathbf{40.72{\pm}3.91}$}\\[1.5pt]
    \cmidrule(lr){1-8}
    \raisebox{3.6pt}{MCXFace} & \raisebox{3.6pt}{$40\!\rightarrow\!40$} & \shortstack{$5.62{\pm}0.37$\\[-1.5pt]$27.24{\pm}1.28$} & \shortstack{$9.95{\pm}4.11$\\[-1.5pt]$41.21{\pm}9.25$} & \shortstack{$6.04{\pm}0.69$\\[-1.5pt]$31.61{\pm}5.06$} & \shortstack{$8.69{\pm}5.47$\\[-1.5pt]$36.60{\pm}13.36$} & \shortstack{$\mathbf{4.86{\pm}0.59}$\\[-1.5pt]$\mathbf{24.76{\pm}2.97}$} & \shortstack{$7.65{\pm}1.62$\\[-1.5pt]$36.11{\pm}4.10$}\\[1.5pt]
    \raisebox{3.6pt}{MCXFace} & \raisebox{3.6pt}{$80\!\rightarrow\!80$} & \shortstack{$\mathbf{4.29{\pm}0.12}$\\[-1.5pt]$\mathbf{21.55{\pm}0.77}$} & \shortstack{$12.04{\pm}2.78$\\[-1.5pt]$48.54{\pm}5.20$} & \shortstack{$4.66{\pm}0.46$\\[-1.5pt]$24.25{\pm}1.80$} & \shortstack{$7.66{\pm}5.40$\\[-1.5pt]$33.06{\pm}16.01$} & \shortstack{$4.93{\pm}1.06$\\[-1.5pt]$25.72{\pm}5.09$} & \shortstack{$7.94{\pm}1.38$\\[-1.5pt]$37.39{\pm}4.88$}\\[1.5pt]
    \raisebox{3.6pt}{MCXFace} & \raisebox{3.6pt}{$160\!\rightarrow\!160$} & \shortstack{$21.65{\pm}2.92$\\[-1.5pt]$62.63{\pm}1.26$} & \shortstack{$11.49{\pm}2.22$\\[-1.5pt]$47.37{\pm}4.86$} & \shortstack{$22.29{\pm}4.02$\\[-1.5pt]$61.57{\pm}1.62$} & \shortstack{$\mathbf{7.48{\pm}2.50}$\\[-1.5pt]$\mathbf{34.75{\pm}10.07}$} & \shortstack{$14.45{\pm}5.67$\\[-1.5pt]$51.16{\pm}8.15$} & \shortstack{$7.94{\pm}1.66$\\[-1.5pt]$37.45{\pm}4.74$}\\[1.5pt]
    \cmidrule(lr){1-8}
    \raisebox{3.6pt}{IIT Delhi} & \raisebox{3.6pt}{$40\!\rightarrow\!40$} & \shortstack{$5.56{\pm}0.45$\\[-1.5pt]$14.25{\pm}0.66$} & \shortstack{$5.09{\pm}0.88$\\[-1.5pt]$12.78{\pm}2.09$} & \shortstack{$7.95{\pm}1.01$\\[-1.5pt]$30.41{\pm}3.38$} & \shortstack{$7.33{\pm}0.58$\\[-1.5pt]$24.06{\pm}3.39$} & \shortstack{$5.14{\pm}0.91$\\[-1.5pt]$13.22{\pm}3.59$} & \shortstack{$\mathbf{3.99{\pm}0.19}$\\[-1.5pt]$\mathbf{9.02{\pm}0.77}$}\\[1.5pt]
    \raisebox{3.6pt}{IIT Delhi} & \raisebox{3.6pt}{$80\!\rightarrow\!80$} & \shortstack{$5.46{\pm}0.52$\\[-1.5pt]$15.95{\pm}1.76$} & \shortstack{$4.52{\pm}0.29$\\[-1.5pt]$11.58{\pm}1.77$} & \shortstack{$7.82{\pm}1.85$\\[-1.5pt]$27.50{\pm}7.94$} & \shortstack{$7.67{\pm}1.34$\\[-1.5pt]$25.82{\pm}5.03$} & \shortstack{$5.55{\pm}0.68$\\[-1.5pt]$14.53{\pm}1.72$} & \shortstack{$\mathbf{3.91{\pm}0.23}$\\[-1.5pt]$\mathbf{8.94{\pm}0.50}$}\\[1.5pt]
    \raisebox{3.6pt}{IIT Delhi} & \raisebox{3.6pt}{$160\!\rightarrow\!160$} & \shortstack{$5.79{\pm}0.11$\\[-1.5pt]$13.82{\pm}0.14$} & \shortstack{$5.51{\pm}1.19$\\[-1.5pt]$14.36{\pm}4.27$} & \shortstack{$5.58{\pm}0.11$\\[-1.5pt]$13.96{\pm}0.88$} & \shortstack{$6.87{\pm}1.06$\\[-1.5pt]$22.34{\pm}7.13$} & \shortstack{$5.55{\pm}0.48$\\[-1.5pt]$15.21{\pm}0.76$} & \shortstack{$\mathbf{4.05{\pm}0.17}$\\[-1.5pt]$\mathbf{9.32{\pm}0.79}$}\\[1.5pt]
    \cmidrule(lr){1-8}
    \raisebox{3.6pt}{MMU1} & \raisebox{3.6pt}{$40\!\rightarrow\!40$} & \shortstack{$10.83{\pm}0.41$\\[-1.5pt]$36.33{\pm}0.99$} & \shortstack{$\mathbf{6.08{\pm}0.06}$\\[-1.5pt]$\mathbf{20.44{\pm}1.95}$} & \shortstack{$11.79{\pm}0.49$\\[-1.5pt]$40.30{\pm}3.02$} & \shortstack{$8.15{\pm}0.79$\\[-1.5pt]$31.04{\pm}12.19$} & \shortstack{$7.48{\pm}0.71$\\[-1.5pt]$27.00{\pm}3.06$} & \shortstack{$\mathbf{6.08{\pm}0.46}$\\[-1.5pt]$21.41{\pm}2.25$}\\[1.5pt]
    \raisebox{3.6pt}{MMU1} & \raisebox{3.6pt}{$80\!\rightarrow\!80$} & \shortstack{$8.63{\pm}0.30$\\[-1.5pt]$32.26{\pm}2.40$} & \shortstack{$\mathbf{5.62{\pm}0.60}$\\[-1.5pt]$22.59{\pm}2.06$} & \shortstack{$9.98{\pm}1.31$\\[-1.5pt]$39.70{\pm}9.43$} & \shortstack{$8.34{\pm}2.52$\\[-1.5pt]$36.00{\pm}12.14$} & \shortstack{$6.70{\pm}1.57$\\[-1.5pt]$24.67{\pm}3.36$} & \shortstack{$5.82{\pm}0.64$\\[-1.5pt]$\mathbf{20.78{\pm}2.41}$}\\[1.5pt]
    \raisebox{3.6pt}{MMU1} & \raisebox{3.6pt}{$160\!\rightarrow\!160$} & \shortstack{$9.21{\pm}0.42$\\[-1.5pt]$33.74{\pm}1.50$} & \shortstack{$\mathbf{5.30{\pm}0.34}$\\[-1.5pt]$21.04{\pm}0.50$} & \shortstack{$8.76{\pm}0.38$\\[-1.5pt]$32.63{\pm}0.83$} & \shortstack{$8.22{\pm}2.90$\\[-1.5pt]$35.78{\pm}21.28$} & \shortstack{$5.51{\pm}0.82$\\[-1.5pt]$\mathbf{19.30{\pm}2.87}$} & \shortstack{$6.48{\pm}0.76$\\[-1.5pt]$22.22{\pm}3.18$}\\[1.5pt]
    \midrule
    \raisebox{3.6pt}{Route macro} & \raisebox{3.6pt}{$40\!\rightarrow\!40$} & \shortstack{$6.00{\pm}0.15$\\[-1.5pt]$23.39{\pm}0.04$} & \shortstack{$6.86{\pm}1.96$\\[-1.5pt]$28.46{\pm}6.86$} & \shortstack{$7.08{\pm}0.19$\\[-1.5pt]$32.37{\pm}3.96$} & \shortstack{$7.36{\pm}2.16$\\[-1.5pt]$35.75{\pm}15.20$} & \shortstack{$\mathbf{4.98{\pm}0.47}$\\[-1.5pt]$\mathbf{19.97{\pm}2.24}$} & \shortstack{$5.99{\pm}1.13$\\[-1.5pt]$25.20{\pm}4.48$}\\[1.5pt]
    \raisebox{3.6pt}{Route macro} & \raisebox{3.6pt}{$80\!\rightarrow\!80$} & \shortstack{$5.06{\pm}0.11$\\[-1.5pt]$\mathbf{20.57{\pm}0.57}$} & \shortstack{$7.50{\pm}1.06$\\[-1.5pt]$34.91{\pm}3.24$} & \shortstack{$6.09{\pm}0.59$\\[-1.5pt]$26.81{\pm}3.69$} & \shortstack{$7.68{\pm}1.88$\\[-1.5pt]$36.38{\pm}4.23$} & \shortstack{$\mathbf{5.03{\pm}0.72}$\\[-1.5pt]$20.92{\pm}2.11$} & \shortstack{$6.12{\pm}1.02$\\[-1.5pt]$26.86{\pm}4.67$}\\[1.5pt]
    \raisebox{3.6pt}{Route macro} & \raisebox{3.6pt}{$160\!\rightarrow\!160$} & \shortstack{$14.76{\pm}1.97$\\[-1.5pt]$50.66{\pm}3.75$} & \shortstack{$8.00{\pm}0.48$\\[-1.5pt]$40.35{\pm}2.03$} & \shortstack{$16.16{\pm}2.94$\\[-1.5pt]$52.40{\pm}3.74$} & \shortstack{$6.84{\pm}1.15$\\[-1.5pt]$32.03{\pm}7.98$} & \shortstack{$9.41{\pm}2.50$\\[-1.5pt]$38.11{\pm}6.70$} & \shortstack{$\mathbf{6.50{\pm}1.13}$\\[-1.5pt]$\mathbf{27.72{\pm}3.48}$}\\[1.5pt]
    \midrule
    \multicolumn{2}{l}{\raisebox{3.6pt}{All-condition macro}} & \shortstack{$\textbf{8.61}{\pm}0.61$\\[-1.5pt]$31.54{\pm}1.21$} & \shortstack{$\textbf{7.45}{\pm}0.62$\\[-1.5pt]$34.58{\pm}1.28$} & \shortstack{$\textbf{9.78}{\pm}0.79$\\[-1.5pt]$37.19{\pm}1.63$} & \shortstack{$\textbf{7.30}{\pm}0.44$\\[-1.5pt]$34.72{\pm}2.08$} & \shortstack{$\textbf{6.47}{\pm}0.98$\\[-1.5pt]${26.33{\pm}2.84}$} & \shortstack{$\mathbf{6.20{\pm}0.87}$\\[-1.5pt]$26.59{\pm}2.78$}\\[1.5pt]
    \bottomrule
  \end{tabular}
\end{table*}

\subsection{RQ2:How can verification be performed effectively across different route image resolutions?}

Table~\ref{tab:reroute-macro} establishes conventional common-route re-extraction as the practical fixed-input baseline and contrasts it with direct cross-model scoring of independent specialists. Direct scoring remains close to chance under both image frequency pool regimes: the routed and full-route macros (average) reach $49.927{\pm}0.114$\% and $50.076{\pm}0.568$\% \textit{EER}.
Presenting every specialist with the complete selected training pool therefore does not align with independently optimized embedding coordinates.

The test protocols use the same fixed source-image pairs for every route combination; the route labels only determine the input rendering and specialist used for each side. 

\begin{table}[]
  \centering
  \caption{Cross-resolution baselines with independent specialists (\%). Both pre-resize source crops are rendered directly for the selected rerouting target. Each cell reports \textit{EER} (top) and \textit{FNMR@FMR} at $1\times10^{-3}$ (bottom) as mean $\pm$ sample standard deviation over three seeds. Bold values compare the two rerouting alternatives.}
  \label{tab:reroute-macro}
  \scriptsize
  \setlength{\tabcolsep}{1.8pt}
  \renewcommand{\arraystretch}{1.0}
  \begin{tabular}{llccc}
    \toprule
    Image pool & Logical pair & Lower route & Higher route & Direct\\
    \midrule
    \raisebox{3.6pt}{Routed} & \raisebox{3.6pt}{$40\!\rightarrow\!80$} & \shortstack{$6.00{\pm}0.15$\\[-1.5pt]$23.39{\pm}0.04$} & \shortstack{$\mathbf{5.06{\pm}0.11}$\\[-1.5pt]$\mathbf{20.57{\pm}0.57}$} & \shortstack{$49.79{\pm}0.96$\\[-1.5pt]$99.91{\pm}0.01$}\\[1.5pt]
    \raisebox{3.6pt}{Routed} & \raisebox{3.6pt}{$40\!\rightarrow\!160$} & \shortstack{$\mathbf{6.00{\pm}0.15}$\\[-1.5pt]$\mathbf{23.39{\pm}0.04}$} & \shortstack{$14.76{\pm}1.97$\\[-1.5pt]$50.66{\pm}3.75$} & \shortstack{$49.76{\pm}0.36$\\[-1.5pt]$99.89{\pm}0.04$}\\[1.5pt]
    \raisebox{3.6pt}{Routed} & \raisebox{3.6pt}{$80\!\rightarrow\!160$} & \shortstack{$\mathbf{5.06{\pm}0.11}$\\[-1.5pt]$\mathbf{20.57{\pm}0.57}$} & \shortstack{$14.76{\pm}1.97$\\[-1.5pt]$50.66{\pm}3.75$} & \shortstack{$50.23{\pm}0.74$\\[-1.5pt]$99.87{\pm}0.13$}\\[1.5pt]
    \addlinespace[1pt]
    \raisebox{3.6pt}{Full} & \raisebox{3.6pt}{$40\!\rightarrow\!80$} & \shortstack{$\mathbf{6.86{\pm}1.96}$\\[-1.5pt]$\mathbf{28.46{\pm}6.86}$} & \shortstack{$7.50{\pm}1.06$\\[-1.5pt]$34.91{\pm}3.24$} & \shortstack{$49.72{\pm}0.47$\\[-1.5pt]$99.90{\pm}0.04$}\\[1.5pt]
    \raisebox{3.6pt}{Full} & \raisebox{3.6pt}{$40\!\rightarrow\!160$} & \shortstack{$\mathbf{6.86{\pm}1.96}$\\[-1.5pt]$\mathbf{28.46{\pm}6.86}$} & \shortstack{$8.00{\pm}0.48$\\[-1.5pt]$40.35{\pm}2.03$} & \shortstack{$50.26{\pm}0.15$\\[-1.5pt]$99.93{\pm}0.01$}\\[1.5pt]
    \raisebox{3.6pt}{Full} & \raisebox{3.6pt}{$80\!\rightarrow\!160$} & \shortstack{$\mathbf{7.50{\pm}1.06}$\\[-1.5pt]$\mathbf{34.91{\pm}3.24}$} & \shortstack{$8.00{\pm}0.48$\\[-1.5pt]$40.35{\pm}2.03$} & \shortstack{$50.25{\pm}1.10$\\[-1.5pt]$99.91{\pm}0.04$}\\[1.5pt]
    \midrule
    \raisebox{3.6pt}{Routed macro} & & \shortstack{$\mathbf{5.69{\pm}0.11}$\\[-1.5pt]$\mathbf{22.45{\pm}0.19}$} & \shortstack{$11.53{\pm}1.31$\\[-1.5pt]$40.63{\pm}2.47$} & \shortstack{$49.93{\pm}0.11$\\[-1.5pt]$99.89{\pm}0.05$}\\[1.5pt]
    \raisebox{3.6pt}{Full macro} & & \shortstack{$\mathbf{7.07{\pm}1.09}$\\[-1.5pt]$\mathbf{30.61{\pm}3.81}$} & \shortstack{$7.83{\pm}0.54$\\[-1.5pt]$38.54{\pm}2.41$} & \shortstack{$50.08{\pm}0.57$\\[-1.5pt]$99.91{\pm}0.01$}\\[1.5pt]
    \bottomrule
  \end{tabular}
\end{table}

Consequently, rerouting a logical $40\rightarrow80$ comparison entirely through route-40 reproduces the corresponding $40\rightarrow40$ evaluation, while rerouting it through route-80 reproduces the $80\rightarrow80$ evaluation. This represents a common-route re-extraction baseline rather than a comparison of image sets with different native resolutions. 

Lower-route rerouting is stronger for both metrics in five of the six route-pair/pool-images conditions and in both resolution-level macros (average). The exception is \(40\rightarrow80\) under routed resolution, where the route-80 specialist has lower matching-route error than route-40. Comparisons involving route-160 strongly favor the lower route because the route-160 specialist trained under routed resolution is the weakest same-resolution specialist in the average dataset. 

Training on full image pools substantially strengthens route-160 and narrows the difference between the alternatives, although the lower route remains stronger for every route pair. 

Table~\ref{tab:cross-eer} compares the conventional lower-route baseline with the proposed direct route scoring mechanisms, \ibsh{} and \shared{}. Routed lower-route rerouting has the best all-condition result at $5.687{\pm}0.112$\% \textit{EER} and $22.449{\pm}0.185$\% \textit{FNMR@FMR}. Full-image pool \shared{} is the strongest directly interoperable system at $6.249{\pm}0.900$\% and $27.512{\pm}3.219$\%.

The condition-level results show where direct interoperability is advantageous. Full-image pool \shared{} outperforms routed lower-route rerouting on all three \textit{IIT Delhi} route pairs. Routed \shared{} is strongest on UFPR and MCXFace for $40\rightarrow80$. In contrast, routed lower-route rerouting dominates \textit{FERET} as well as the $40\rightarrow160$ and $80\rightarrow160$ comparisons on \textit{UFPR} and \textit{MCXFace}. 

Training on route image pools in lower-route rerouting is strongest across all three \textit{MMU1} route pairs. Shared class centers alone provide partial interoperability.

The \ibsh{} reduces cross-resolution error far below the independent direct null control, but both \ibsh{} exposure variants remain behind the shared trunk and common-route re-extraction in the all-condition macros.

The practical conclusion depends on the available evidence at comparison time. When source images can be reprocessed, lower-route rerouting represents conventional fixed-input practice and remains the strongest strategy on average in these experiments. 

The shared trunk nevertheless surpasses this baseline in several dataset--route conditions. It also provides the strongest interoperability when previously generated route-native templates must be scored directly, avoiding mandatory common-route re-extraction at a measurable average accuracy.

\begin{table*}[t]
  \centering
  \caption{Cross-resolution verification strategies (\%) for all dataset--route conditions. Independent columns re-extract both observations with the lower-route specialist; \ibsh{} and \shared{} columns directly compare route-native embeddings. Each cell reports \textit{EER} (top) and \textit{FNMR@FMR} at $1\times10^{-3}$ (bottom) as mean $\pm$ sample standard deviation over three seeds. Bold values are row-wise best for each metric.}
  \label{tab:cross-eer}
  \scriptsize
  \setlength{\tabcolsep}{1.5pt}
  \renewcommand{\arraystretch}{1.0}
  \begin{tabular}{@{}ll@{\hspace{2.5pt}\vrule width 0.2pt\hspace{2.5pt}}cc@{\hspace{2.5pt}\vrule width 0.2pt\hspace{2.5pt}}cc@{\hspace{2.5pt}\vrule width 0.2pt\hspace{2.5pt}}cc@{}}
    \toprule
    & & \multicolumn{2}{c}{\ind{} lower route} & \multicolumn{2}{c}{\ibsh{} direct} & \multicolumn{2}{c}{\shared{} direct}\\
    \cmidrule(lr){3-4}\cmidrule(lr){5-6}\cmidrule(lr){7-8}
    Dataset & Route & 1/3 & 3/3 & 1/3 & 3/3 & 1/3 & 3/3\\
    \midrule
    \raisebox{3.6pt}{FERET} & \raisebox{3.6pt}{$40\!\rightarrow\!80$} & \shortstack{$\mathbf{1.30{\pm}0.06}$\\[-1.5pt]$\mathbf{6.81{\pm}0.65}$} & \shortstack{$5.50{\pm}3.57$\\[-1.5pt]$30.29{\pm}16.43$} & \shortstack{$2.03{\pm}0.74$\\[-1.5pt]$7.87{\pm}1.99$} & \shortstack{$7.66{\pm}4.44$\\[-1.5pt]$25.76{\pm}13.10$} & \shortstack{$1.76{\pm}0.15$\\[-1.5pt]$8.23{\pm}1.39$} & \shortstack{$4.46{\pm}1.76$\\[-1.5pt]$23.51{\pm}8.53$}\\[1.5pt]
    \raisebox{3.6pt}{FERET} & \raisebox{3.6pt}{$40\!\rightarrow\!160$} & \shortstack{$\mathbf{1.30{\pm}0.06}$\\[-1.5pt]$\mathbf{6.81{\pm}0.65}$} & \shortstack{$5.50{\pm}3.57$\\[-1.5pt]$30.29{\pm}16.43$} & \shortstack{$20.05{\pm}7.34$\\[-1.5pt]$76.06{\pm}17.40$} & \shortstack{$5.71{\pm}4.00$\\[-1.5pt]$19.22{\pm}8.11$} & \shortstack{$8.30{\pm}4.29$\\[-1.5pt]$39.67{\pm}20.64$} & \shortstack{$4.82{\pm}2.53$\\[-1.5pt]$25.41{\pm}12.08$}\\[1.5pt]
    \raisebox{3.6pt}{FERET} & \raisebox{3.6pt}{$80\!\rightarrow\!160$} & \shortstack{$\mathbf{1.18{\pm}0.19}$\\[-1.5pt]$\mathbf{5.30{\pm}0.87}$} & \shortstack{$7.11{\pm}2.35$\\[-1.5pt]$53.46{\pm}12.12$} & \shortstack{$19.68{\pm}7.68$\\[-1.5pt]$75.32{\pm}17.80$} & \shortstack{$4.93{\pm}4.95$\\[-1.5pt]$19.16{\pm}16.96$} & \shortstack{$7.84{\pm}3.84$\\[-1.5pt]$40.96{\pm}20.95$} & \shortstack{$4.95{\pm}1.74$\\[-1.5pt]$26.95{\pm}7.56$}\\[1.5pt]
    \cmidrule(lr){1-8}
    \raisebox{3.6pt}{UFPR} & \raisebox{3.6pt}{$40\!\rightarrow\!80$} & \shortstack{$6.69{\pm}0.01$\\[-1.5pt]$32.32{\pm}0.51$} & \shortstack{$7.70{\pm}1.29$\\[-1.5pt]$37.59{\pm}5.72$} & \shortstack{$8.25{\pm}0.29$\\[-1.5pt]$35.67{\pm}1.25$} & \shortstack{$10.56{\pm}1.59$\\[-1.5pt]$41.43{\pm}2.71$} & \shortstack{$\mathbf{6.19{\pm}0.34}$\\[-1.5pt]$\mathbf{30.03{\pm}1.37}$} & \shortstack{$8.12{\pm}0.72$\\[-1.5pt]$39.73{\pm}2.61$}\\[1.5pt]
    \raisebox{3.6pt}{UFPR} & \raisebox{3.6pt}{$40\!\rightarrow\!160$} & \shortstack{$\mathbf{6.69{\pm}0.01}$\\[-1.5pt]$\mathbf{32.32{\pm}0.51}$} & \shortstack{$7.70{\pm}1.29$\\[-1.5pt]$37.59{\pm}5.72$} & \shortstack{$22.50{\pm}5.26$\\[-1.5pt]$77.75{\pm}6.97$} & \shortstack{$9.36{\pm}1.36$\\[-1.5pt]$39.15{\pm}4.29$} & \shortstack{$10.20{\pm}1.57$\\[-1.5pt]$46.24{\pm}5.10$} & \shortstack{$8.40{\pm}0.92$\\[-1.5pt]$40.60{\pm}4.32$}\\[1.5pt]
    \raisebox{3.6pt}{UFPR} & \raisebox{3.6pt}{$80\!\rightarrow\!160$} & \shortstack{$\mathbf{5.75{\pm}0.23}$\\[-1.5pt]$\mathbf{27.77{\pm}1.43}$} & \shortstack{$8.19{\pm}0.77$\\[-1.5pt]$38.41{\pm}3.05$} & \shortstack{$21.60{\pm}5.24$\\[-1.5pt]$76.18{\pm}7.22$} & \shortstack{$10.40{\pm}0.60$\\[-1.5pt]$42.67{\pm}1.41$} & \shortstack{$9.30{\pm}1.24$\\[-1.5pt]$43.72{\pm}5.12$} & \shortstack{$8.30{\pm}0.55$\\[-1.5pt]$41.06{\pm}2.25$}\\[1.5pt]
    \cmidrule(lr){1-8}
    \raisebox{3.6pt}{MCXFace} & \raisebox{3.6pt}{$40\!\rightarrow\!80$} & \shortstack{$5.62{\pm}0.37$\\[-1.5pt]$27.24{\pm}1.28$} & \shortstack{$9.95{\pm}4.11$\\[-1.5pt]$41.21{\pm}9.25$} & \shortstack{$5.31{\pm}0.78$\\[-1.5pt]$26.50{\pm}2.52$} & \shortstack{$9.40{\pm}4.06$\\[-1.5pt]$40.31{\pm}12.56$} & \shortstack{$\mathbf{4.57{\pm}0.65}$\\[-1.5pt]$\mathbf{23.93{\pm}2.57}$} & \shortstack{$7.74{\pm}1.44$\\[-1.5pt]$36.23{\pm}2.68$}\\[1.5pt]
    \raisebox{3.6pt}{MCXFace} & \raisebox{3.6pt}{$40\!\rightarrow\!160$} & \shortstack{$\mathbf{5.62{\pm}0.37}$\\[-1.5pt]$\mathbf{27.24{\pm}1.28}$} & \shortstack{$9.95{\pm}4.11$\\[-1.5pt]$41.21{\pm}9.25$} & \shortstack{$15.60{\pm}3.30$\\[-1.5pt]$55.90{\pm}2.96$} & \shortstack{$9.03{\pm}4.11$\\[-1.5pt]$37.61{\pm}11.12$} & \shortstack{$8.67{\pm}2.82$\\[-1.5pt]$38.51{\pm}9.01$} & \shortstack{$7.71{\pm}1.64$\\[-1.5pt]$36.16{\pm}5.19$}\\[1.5pt]
    \raisebox{3.6pt}{MCXFace} & \raisebox{3.6pt}{$80\!\rightarrow\!160$} & \shortstack{$\mathbf{4.29{\pm}0.12}$\\[-1.5pt]$\mathbf{21.55{\pm}0.77}$} & \shortstack{$12.04{\pm}2.78$\\[-1.5pt]$48.54{\pm}5.20$} & \shortstack{$14.67{\pm}3.62$\\[-1.5pt]$54.44{\pm}3.45$} & \shortstack{$8.62{\pm}5.87$\\[-1.5pt]$36.78{\pm}20.47$} & \shortstack{$8.81{\pm}3.17$\\[-1.5pt]$40.44{\pm}11.92$} & \shortstack{$8.00{\pm}1.10$\\[-1.5pt]$38.05{\pm}3.56$}\\[1.5pt]
    \cmidrule(lr){1-8}
    \raisebox{3.6pt}{IIT Delhi} & \raisebox{3.6pt}{$40\!\rightarrow\!80$} & \shortstack{$5.56{\pm}0.45$\\[-1.5pt]$14.25{\pm}0.66$} & \shortstack{$5.09{\pm}0.88$\\[-1.5pt]$12.78{\pm}2.09$} & \shortstack{$25.91{\pm}3.43$\\[-1.5pt]$84.13{\pm}1.82$} & \shortstack{$29.41{\pm}6.88$\\[-1.5pt]$90.39{\pm}1.17$} & \shortstack{$7.04{\pm}1.21$\\[-1.5pt]$39.63{\pm}11.12$} & \shortstack{$\mathbf{4.04{\pm}0.15}$\\[-1.5pt]$\mathbf{10.01{\pm}0.71}$}\\[1.5pt]
    \raisebox{3.6pt}{IIT Delhi} & \raisebox{3.6pt}{$40\!\rightarrow\!160$} & \shortstack{$5.56{\pm}0.45$\\[-1.5pt]$14.25{\pm}0.66$} & \shortstack{$5.09{\pm}0.88$\\[-1.5pt]$12.78{\pm}2.09$} & \shortstack{$25.05{\pm}1.68$\\[-1.5pt]$88.39{\pm}1.23$} & \shortstack{$26.90{\pm}12.93$\\[-1.5pt]$82.83{\pm}5.52$} & \shortstack{$10.49{\pm}3.14$\\[-1.5pt]$76.46{\pm}6.27$} & \shortstack{$\mathbf{4.27{\pm}0.20}$\\[-1.5pt]$\mathbf{11.94{\pm}1.84}$}\\[1.5pt]
    \raisebox{3.6pt}{IIT Delhi} & \raisebox{3.6pt}{$80\!\rightarrow\!160$} & \shortstack{$5.46{\pm}0.52$\\[-1.5pt]$15.95{\pm}1.76$} & \shortstack{$4.52{\pm}0.29$\\[-1.5pt]$11.58{\pm}1.77$} & \shortstack{$20.22{\pm}5.68$\\[-1.5pt]$84.14{\pm}1.55$} & \shortstack{$39.02{\pm}4.50$\\[-1.5pt]$90.88{\pm}3.03$} & \shortstack{$7.23{\pm}1.07$\\[-1.5pt]$39.17{\pm}7.93$} & \shortstack{$\mathbf{4.06{\pm}0.13}$\\[-1.5pt]$\mathbf{10.14{\pm}1.24}$}\\[1.5pt]
    \cmidrule(lr){1-8}
    \raisebox{3.6pt}{MMU1} & \raisebox{3.6pt}{$40\!\rightarrow\!80$} & \shortstack{$10.83{\pm}0.41$\\[-1.5pt]$36.33{\pm}0.99$} & \shortstack{$\mathbf{6.08{\pm}0.06}$\\[-1.5pt]$\mathbf{20.44{\pm}1.95}$} & \shortstack{$21.84{\pm}2.70$\\[-1.5pt]$75.33{\pm}9.68$} & \shortstack{$20.00{\pm}10.38$\\[-1.5pt]$65.33{\pm}14.15$} & \shortstack{$7.82{\pm}0.74$\\[-1.5pt]$32.07{\pm}2.83$} & \shortstack{$6.24{\pm}0.49$\\[-1.5pt]$23.93{\pm}3.84$}\\[1.5pt]
    \raisebox{3.6pt}{MMU1} & \raisebox{3.6pt}{$40\!\rightarrow\!160$} & \shortstack{$10.83{\pm}0.41$\\[-1.5pt]$36.33{\pm}0.99$} & \shortstack{$\mathbf{6.08{\pm}0.06}$\\[-1.5pt]$\mathbf{20.44{\pm}1.95}$} & \shortstack{$20.58{\pm}0.79$\\[-1.5pt]$87.00{\pm}1.56$} & \shortstack{$17.96{\pm}6.63$\\[-1.5pt]$59.96{\pm}16.10$} & \shortstack{$8.37{\pm}1.01$\\[-1.5pt]$37.48{\pm}5.25$} & \shortstack{$6.33{\pm}0.44$\\[-1.5pt]$25.11{\pm}3.19$}\\[1.5pt]
    \raisebox{3.6pt}{MMU1} & \raisebox{3.6pt}{$80\!\rightarrow\!160$} & \shortstack{$8.63{\pm}0.30$\\[-1.5pt]$32.26{\pm}2.40$} & \shortstack{$\mathbf{5.62{\pm}0.60}$\\[-1.5pt]$\mathbf{22.59{\pm}2.06}$} & \shortstack{$17.82{\pm}2.48$\\[-1.5pt]$79.19{\pm}7.27$} & \shortstack{$20.57{\pm}9.12$\\[-1.5pt]$60.15{\pm}14.70$} & \shortstack{$6.76{\pm}1.31$\\[-1.5pt]$27.70{\pm}4.10$} & \shortstack{$6.29{\pm}0.77$\\[-1.5pt]$23.85{\pm}3.05$}\\[1.5pt]
    \midrule
    \raisebox{3.6pt}{Route macro} & \raisebox{3.6pt}{$40\!\rightarrow\!80$} & \shortstack{$6.00{\pm}0.15$\\[-1.5pt]$\mathbf{23.39{\pm}0.04}$} & \shortstack{$6.86{\pm}1.96$\\[-1.5pt]$28.46{\pm}6.86$} & \shortstack{$12.67{\pm}1.04$\\[-1.5pt]$45.90{\pm}1.74$} & \shortstack{$15.40{\pm}4.78$\\[-1.5pt]$52.64{\pm}5.58$} & \shortstack{$\mathbf{5.47{\pm}0.56}$\\[-1.5pt]$26.78{\pm}3.27$} & \shortstack{$6.12{\pm}0.88$\\[-1.5pt]$26.68{\pm}3.12$}\\[1.5pt]
    \raisebox{3.6pt}{Route macro} & \raisebox{3.6pt}{$40\!\rightarrow\!160$} & \shortstack{$\mathbf{6.00{\pm}0.15}$\\[-1.5pt]$\mathbf{23.39{\pm}0.04}$} & \shortstack{$6.86{\pm}1.96$\\[-1.5pt]$28.46{\pm}6.86$} & \shortstack{$20.76{\pm}2.95$\\[-1.5pt]$77.02{\pm}5.15$} & \shortstack{$13.79{\pm}5.06$\\[-1.5pt]$47.76{\pm}5.82$} & \shortstack{$9.21{\pm}1.92$\\[-1.5pt]$47.67{\pm}5.41$} & \shortstack{$6.31{\pm}1.11$\\[-1.5pt]$27.84{\pm}4.58$}\\[1.5pt]
    \raisebox{3.6pt}{Route macro} & \raisebox{3.6pt}{$80\!\rightarrow\!160$} & \shortstack{$\mathbf{5.06{\pm}0.11}$\\[-1.5pt]$\mathbf{20.57{\pm}0.57}$} & \shortstack{$7.50{\pm}1.06$\\[-1.5pt]$34.91{\pm}3.24$} & \shortstack{$18.80{\pm}1.69$\\[-1.5pt]$73.85{\pm}4.04$} & \shortstack{$16.71{\pm}1.66$\\[-1.5pt]$49.93{\pm}5.17$} & \shortstack{$7.99{\pm}1.42$\\[-1.5pt]$38.40{\pm}5.23$} & \shortstack{$6.32{\pm}0.83$\\[-1.5pt]$28.01{\pm}2.87$}\\[1.5pt]
    \midrule
    \multicolumn{2}{l}{\raisebox{3.6pt}{All-condition macro}} & \shortstack{$\mathbf{5.69{\pm}0.11}$\\[-1.5pt]$\mathbf{22.45{\pm}0.19}$} & \shortstack{$7.07{\pm}1.09$\\[-1.5pt]$30.61{\pm}3.81$} & \shortstack{$17.41{\pm}1.18$\\[-1.5pt]$65.59{\pm}2.51$} & \shortstack{$15.30{\pm}2.84$\\[-1.5pt]$50.11{\pm}2.24$} & \shortstack{$7.56{\pm}1.16$\\[-1.5pt]$37.62{\pm}3.18$} & \shortstack{$6.25{\pm}0.90$\\[-1.5pt]$27.51{\pm}3.22$}\\[1.5pt]
    \bottomrule
  \end{tabular}
\end{table*}

\subsection{RQ3:How does training-data routes influence the specialization and generalization of resolution-specific models?}

The route macros (average) in Table~\ref{tab:same-eer} show a coherent image pool response across all three architectures. Routed specialist pool images yields lower \textit{EER} and \textit{FNMR@FMR} on matching routes $40$ and $80$, while training on full image pools yields lower errors on route $160$. Presenting every selected source to every route therefore does not provide a uniform same-resolution improvement.

Table~\ref{tab:shared-exposure-delta} resolves this effect by dataset for the shared-trunk system. \textit{FERET}, \textit{UFPR} and \textit{MCXFace} exhibit the same pattern: routed specialist improves routes $40$ and $80$, while training on full image pools improves route-160. 

\textit{The MMU1} dataset shows the complementary iris-oriented pattern, with training on full image pools improving routes $40$ and $80$ but routed specialist retaining the advantage on route-160.

\textit{IIT Delhi} dataset is the exception because full exposure improves all three routes. The signs agree between \textit{EER} and \textit{FNMR@FMR} in every dataset--route condition.

\begin{table}[t]
  \centering
  \caption{Shared-trunk response to full instead of routed pool image in percentage points. Each cell reports $\Delta\textit{EER}$ (top) and $\Delta\textit{FNMR@FMR}$ at $1\times10^{-3}$ (bottom), where $\Delta m=m_{\mathrm{full}}-m_{\mathrm{routed}}$. Positive values favor specialist routed; negative values favor on training on full image pools. Differences are computed from the unrounded three-seed means.}
  \label{tab:shared-exposure-delta}
  \scriptsize
  \setlength{\tabcolsep}{4.2pt}
  \renewcommand{\arraystretch}{1.0}
  \begin{tabular}{lccc}
    \toprule
    Dataset & Route 40 & Route 80 & Route 160\\
    \midrule
    \raisebox{3.6pt}{FERET} & \shortstack{$+2.90$\\[-1.5pt]$+16.03$} & \shortstack{$+3.05$\\[-1.5pt]$+17.64$} & \shortstack{$-5.93$\\[-1.5pt]$-25.21$}\\[1.5pt]
    \raisebox{3.6pt}{UFPR} & \shortstack{$+1.89$\\[-1.5pt]$+8.56$} & \shortstack{$+1.93$\\[-1.5pt]$+9.84$} & \shortstack{$-1.62$\\[-1.5pt]$-10.02$}\\[1.5pt]
    \raisebox{3.6pt}{MCXFace} & \shortstack{$+2.78$\\[-1.5pt]$+11.36$} & \shortstack{$+3.01$\\[-1.5pt]$+11.67$} & \shortstack{$-6.51$\\[-1.5pt]$-13.71$}\\[1.5pt]
    \raisebox{3.6pt}{IIT Delhi} & \shortstack{$-1.15$\\[-1.5pt]$-4.20$} & \shortstack{$-1.64$\\[-1.5pt]$-5.59$} & \shortstack{$-1.50$\\[-1.5pt]$-5.90$}\\[1.5pt]
    \raisebox{3.6pt}{MMU1} & \shortstack{$-1.40$\\[-1.5pt]$-5.59$} & \shortstack{$-0.87$\\[-1.5pt]$-3.89$} & \shortstack{$+0.96$\\[-1.5pt]$+2.93$}\\[1.5pt]
    \bottomrule
  \end{tabular}
\end{table}

Training on full image pools also contracts the within-dataset \textit{EER} range across routes in all five datasets. The range decreases from $10.12$ to $1.30$ points for \textit{FERET}, from $4.09$ to $0.54$ for \textit{UFPR}, from $9.58$ to $0.29$ for \textit{MCXFace}, from $0.41$ to $0.14$ for \textit{IIT Delhi} and from $1.97$ to $0.65$ for \textit{MMU1}. 

Routed exposure therefore creates sharper operational route specialists. Training on full image pools homogenizes their performance, improving routes that are poorly supported by their routed training pool while weakening several of the strongest matched specialists.

The same image pool response explains the common-route results in Table~\ref{tab:reroute-macro}. Across the three route pairs, lower-route rerouting selects route-40 twice and route-80 once. Higher-route rerouting selects route-80 once and route-160 twice. 

Under routed-specific image pool, the strong route-40 and route-80 specialists produce a $5.687{\pm}0.112$\% lower-route macro-\textit{EER}, while the weak route-160 specialist raises the higher-route result to $11.525{\pm}1.314$\%. 

Full exposure improves the route-160 macro-\textit{EER} from $14.757{\pm}1.974$\% to $8.004{\pm}0.478$\%, but route-40 changes from $6.001{\pm}0.148$\% to $6.861{\pm}1.963$\% and route 80 from $5.060{\pm}0.109$\% to $7.496{\pm}1.057$\%. Consequently, higher-route rerouting improves under full exposure while lower-route rerouting becomes weaker.

Training on full image pools also improves direct cross-resolution scoring: the \ibsh{} macro-\textit{EER} decreases from $17.411{\pm}1.182$\% to $15.301{\pm}2.839$\%, while the \shared{} result decreases from $7.556{\pm}1.159$\% to $6.249{\pm}0.900$\%. Route-specific resolution therefore provides the strongest matched specialists in several data-rich conditions, whereas training on full image pools trades these peaks for more uniform route behavior and stronger direct interoperability.

\section{Discussion and Limitations}
\label{sec:discussion}

The three research questions reveal a trade-off between same-resolution specialization and direct cross-resolution interoperability. No architecture or image pool regime dominates every matching-route condition. Conventional lower-route re-extraction remains stronger in aggregate when source observations can be reprocessed through one fixed-resolution specialist. Shared class centers provide partial cross-route alignment, while the \textit{\textbf{shared trunk produces the strongest}} directly interoperable templates on average and surpasses the conventional baseline in selected conditions.

\subsection{Implications for model selection}

Coupling should be treated as a deployment design axis rather than as a free accuracy improvement. When source images can be reprocessed, lower-route rerouting represents the conventional fixed-input baseline and remains the strongest strategy on average in the present tests. 

The shared trunk nevertheless surpasses this baseline in selected dataset route conditions and enables direct comparison of route-native templates without common-route re-extraction. 

\textit{IBSH} occupies an intermediate point: it retains complete route-specific networks and already produces a common score space, but it remains behind the conventional baseline and the shared trunk in aggregate cross-resolution performance.

Training on the full image pool is a ,second deployment design axis. Routed specialist resolution produces sharper route-dependent performance profiles, while training on the full image pool contracts the differences among routes. 

For the shared trunk, routed exposure improves routes-40 and 80 on \textit{FERET}, \textit{UFPR} and \textit{MCXFace} and retains the route-160 advantage on \textit{MMU1}. 

Training on the full image pool instead strengthens under-supported routes and provides the best results on all \textit{IIT Delhi} routes. 

Additional heterogeneous data can therefore suppress a matched specialist advantage even while improving robustness elsewhere. The highest tensor resolution is likewise not universally optimal. These findings suggest that more is not always better in a straightforward way. Route capacity, source content, and image pool composition must be considered together.

\subsection{Limitations}

Training on the full image pool comparison is not compute-matched. Routed training processes $27.64$ million route views over the complete schedule, whereas training on the full image pool processes $75.57$ million. 

Full resolution changes the number of route views and the diversity seen by each specialist. The comparison therefore measures the practical effect of the two policies rather than isolating unique data volume, repeated full image pool, or specialization as separate causal factors.

The architecture comparisons do not isolate every mechanism. \ibsh{} changes joint optimization, effective head batch and \textit{AdaFace} norm statistics in addition to sharing class centers. 

A synchronized separate-head control would be required to attribute alignment solely to the common centers. The \ibsh{}-to-\shared{} comparison also changes parameter sharing and active computation, so it remains a system comparison rather than an isolated trunk ablation. Only one AdaFace-based backbone family, one embedding dimension, three fixed input routes and one shared-trunk interface are evaluated.

The experiments deliberately retain a common AdaFace-style training recipe rather than tuning each architecture independently. This choice prioritizes controlled comparability over the maximum attainable performance of any individual system. Because coupling and specialist exposure alter optimization dynamics substantially, architecture-specific learning schedules, regularization and augmentation policies could improve absolute performance or change the relative ranking.

Dataset, spectrum, sensor, crop geometry, resolution and visible ocular anatomy co-vary in the available data. The observed operational route specialization therefore cannot be attributed to tensor resolution alone. Isolating a pure resolution effect would require a controlled single-domain experiment with matched identities, source images and exposure across routes. 


Source-dataset separation does not guarantee global identity separation from independently collected or internet-scraped training sets. Cross-route verification demonstrates compatibility under the heterogeneous test protocols, but it does not independently prove native \textit{VIS-periocular} to NIR-iris interoperability.


\section{Conclusion}
\label{sec:conclusion}

This study demonstrates that multi-resolution ocular recognition can be effectively optimized by matching model training strategies to the available image evidence and deployment requirements. Route-specific training enables strong specialization: routes 40 and 80 achieve particularly high performance on their assigned data, especially for FERET, \textit{UFPR}, and \textit{MCXFace}. By contrast, training on the full image pool promotes more consistent performance across routes, reducing specialist peaks while improving overall balance. Route 160 benefits especially from this broader training strategy, achieving stronger performance across all three architectures, while the route-restricted shared-trunk model maintains an advantage on \textit{MMU1}.

The results also highlight complementary strengths among the model configurations. The \textit{shared-trunk model} trained on the full image pool achieves the lowest same-resolution macro-EER, while its route-restricted counterpart achieves the lowest same-resolution FNMR@FMR. These findings show that model resolution, training-pool composition, and parameter sharing can be selected strategically to target different verification objectives.

The study also shows that shared representations can improve template interoperability. Although independently trained specialist embeddings are not directly interoperable, shared class centers provide partial alignment, and the shared trunk produces the most interoperable templates on average. When source images can be reprocessed, the conventional lower-route re-extraction baseline remains the strongest overall approach; however, the shared trunk performs better under selected conditions while enabling direct interoperability between route-native templates.

Overall, the findings establish that route specialization and template interoperability are complementary but distinct deployment objectives. Rather than relying on a single universally optimal configuration, effective multi-resolution ocular recognition is best achieved by aligning image-pool design, tensor resolution, parameter sharing, and template-processing workflows with the available image evidence and operational requirements.

\section*{Acknowledgment}

This work was partially supported by EU Horizon under G.A. CarMen (101168325), and the German Federal Ministry of Education and Research and the Hessian Ministry of Higher Education, Research, Science and the Arts within their joint support of the National Research Center for Applied Cybersecurity ATHENE.

\bibliographystyle{IEEEtran}
\bibliography{references}

\begin{IEEEbiography}[{\includegraphics[width=0.9in,height=1.25in,clip,keepaspectratio]{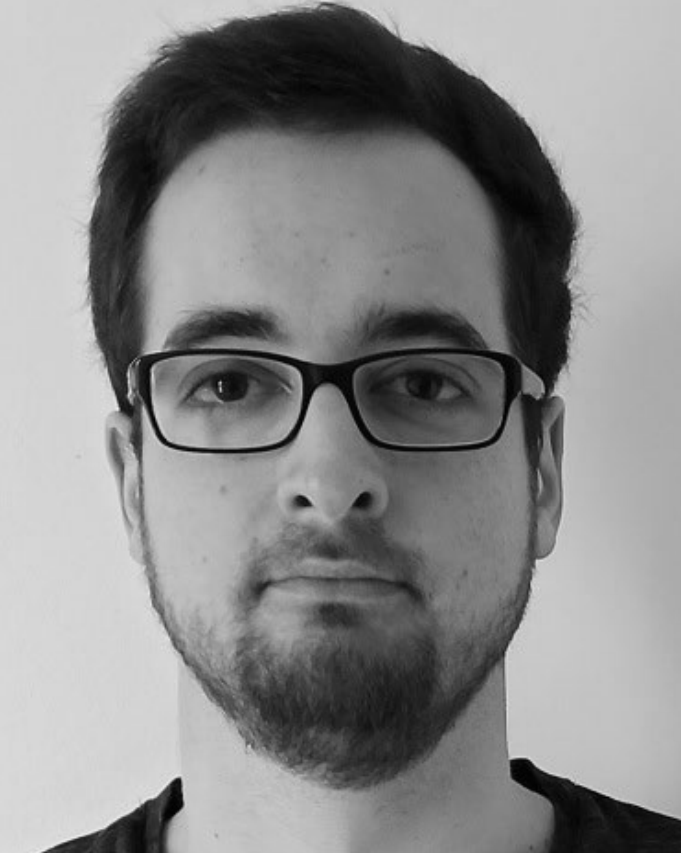}}]{Fabian Stockhardt}
is a research member of the Darmstadt-based da/sec research group. He received the LL.B. and B.Sc. degrees from Hochschule Darmstadt. His main research interests include computer vision, biometrics, privacy, and deep learning systems.
\end{IEEEbiography}
\vspace{-10pt}

\begin{IEEEbiography}[{\includegraphics[width=0.9in,height=1.25in,clip,keepaspectratio]{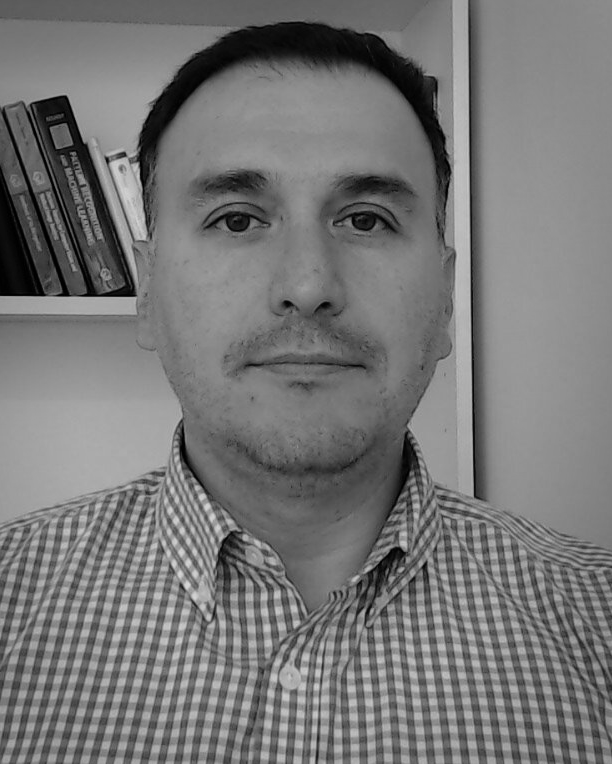}}]{Juan Tapia}
is a Senior Researcher at Hochschule Darmstadt (h-da) in Germany, where he leads EU-funded projects like EINSTEIN, iMARS, and CARMEN in biometrics and digital forensics. With over 15 years of experience in artificial intelligence, computer vision, and identity security, he is an entrepreneur and leading expert in Iris and Face Recognition, Presentation Attack Detection (PAD) for ID documents, Fitness-For-Duty evaluation (FFD), and generative AI for biometrics. He holds a PhD in Electrical Engineering from the Universidad de Chile (2016), an MSc (2012), and a BSc in Electronics Engineering from Universidad Mayor (2004). He was awarded the Best PhD Thesis Award in 2016. Professor at several Universities in Chile in the computer science department. Dr. Tapia is an active reviewer for top-tier journals and conferences and serves as an Associate Editor for the IET Image Processing Journal. He is also a member of the German DIN on behalf of the ISO/IEC SC 37 on biometrics. His research focuses on foundation models (ViT, GANs, diffusion models), deep learning, and pattern recognition applied to biometrics, morphing, and Presentation Attack Detection on iris, faces, and ID cards. He is passionate about innovating in security and protecting people and organizations through AI-driven solutions. 
\end{IEEEbiography}
\vspace{-6pt}

\begin{IEEEbiography}[{\includegraphics[width=0.9in,height=1.25in,clip,keepaspectratio]{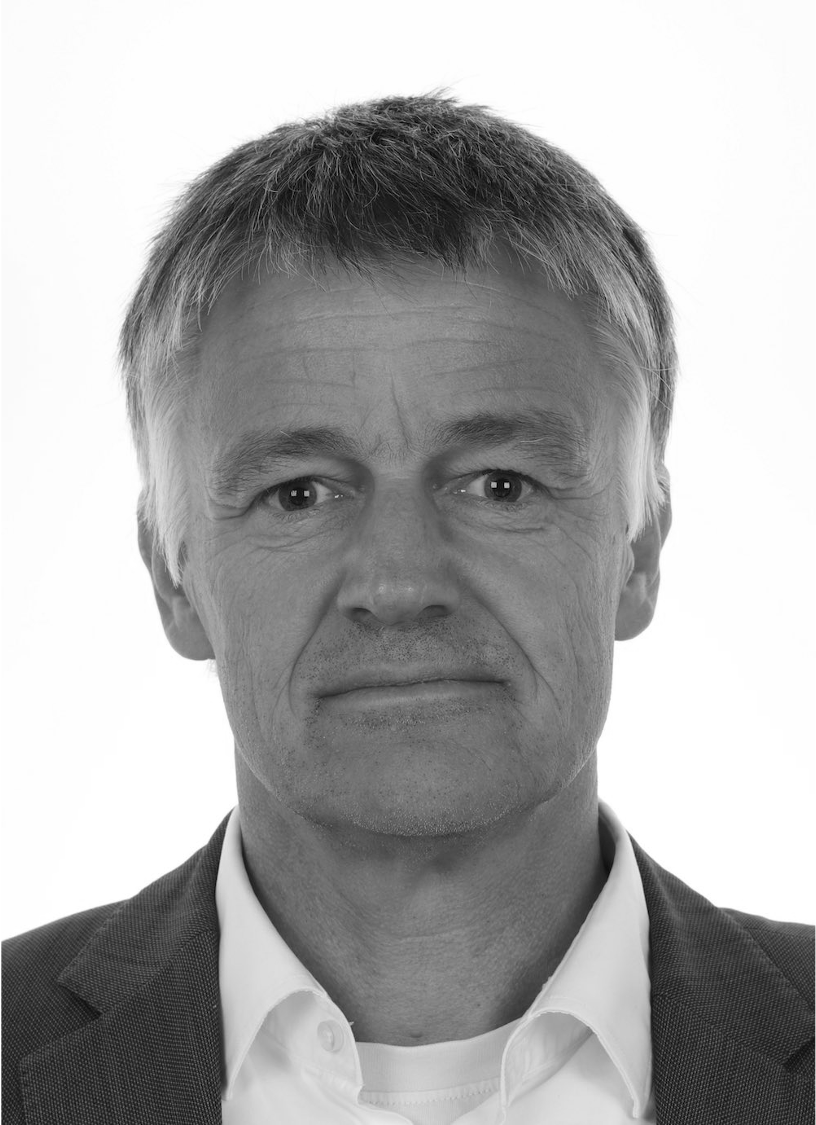}}]{Christoph Busch}
is a member of the Department of Information Security and Communication Technology (IIK) at the Norwegian University of Science and Technology (NTNU), Norway. He holds a joint appointment with the computer science faculty at Hochschule Darmstadt (HDA), Germany. On behalf of the German BSI, he has been the coordinator for the project series BioIS, BioFace, BioFinger, BioKeyS Pilot-DB, KBEinweg and NFIQ2.0. In the European research program, he was the initiator of the Integrated Project 3D-Face, FIDELITY and iMARS. Further, he was/is a partner in the projects TURBINE, BEST Network, ORIGINS, INGRESS, PIDaaS, SOTAMD, RESPECT and TReSPAsS. He is also the principal investigator at the German National Research Centre for Applied Cybersecurity (ATHENE). Moreover, Christoph Busch is a co-founder and board member of the European Association for Biometrics (www.eab.org), which was established in 2011 and comprises more than 200 institutional members. Christoph co-authored more than 700 technical papers and has been a speaker at international conferences. Christoph Busch is a member of the German Gesellschaft fur Informatik (GI) and the International Association of Pattern Recognition (IAPR), as well as an IEEE Fellow.
\end{IEEEbiography}

\vfill

\end{document}